\documentclass{article}

\PassOptionsToPackage{square, numbers, comma, sort&compress}{natbib}

\usepackage[final]{neurips_2025}

\usepackage[T1]{fontenc}
\usepackage{hyperref}
\usepackage{url}
\usepackage{booktabs}
\usepackage{amsfonts}
\usepackage{nicefrac}
\usepackage{microtype}
\usepackage{xspace}
\usepackage{xpatch}
\usepackage{multirow}
\usepackage{amsmath}
\usepackage{graphicx}
\usepackage{bm}
\usepackage{enumitem}
\usepackage{algpseudocode}
\usepackage{algorithm}
\usepackage{subfig}
\usepackage{subfloat}
\usepackage{amssymb} 
\usepackage{enumitem}
\usepackage{soul}
\usepackage{wrapfig}
\usepackage{marvosym}
\usepackage[dvipsnames]{xcolor}
\usepackage{colortbl}

\definecolor{dark_green}{rgb}{0, 0.5, 0}
\definecolor{dark_red}{rgb}{0.8, 0.2, 0.2}
\definecolor{soft_red}{rgb}{1.0, 0.4, 0.4}

\definecolor{citecolor}{HTML}{0071bc}
\definecolor{paleplum}{rgb}{0.8, 0.6, 0.8}
\hypersetup{
  colorlinks,
  citecolor=citecolor,
  linkcolor=red
}

\usepackage[capitalise]{cleveref}
\crefname{figure}{Fig.}{Figs.}
\crefname{table}{Tab.}{Tabs.}
\crefname{section}{Sec.}{Secs.}
\crefname{subsection}{Sec.}{Secs.}
\crefname{equation}{Eq.}{Eqs.}

\usepackage{xpatch}
\makeatletter
\xapptocmd{\NAT@bibsetnum}{\setlength{\leftmargin}{0pt}\setlength{\itemindent}{\labelwidth}\addtolength{\itemindent}{\labelsep}}{}{}
\makeatother

\def\eg{\emph{e.g.}\xspace} 
\def\ie{\emph{i.e.}\xspace}

\def\aka{\emph{a.k.a.}\xspace}

\newcommand{\boldstart}[1]{\vspace{0.1in}\noindent\textbf{#1}}

\newcommand{\method}{ZPressor\xspace}

\newcommand\rurl[1]{%
\href{https://#1}{\nolinkurl{#1}}%
}

\def\mZ{{\mathcal Z}}
\def\mX{{\mathcal X}}
\def\mY{{\mathcal Y}}

\title{\method: Bottleneck-Aware Compression for \\ Scalable Feed-Forward 3DGS}

\author{
    Weijie Wang\textsuperscript{1} \quad
    Donny Y. Chen\textsuperscript{2} \quad
    Zeyu Zhang\textsuperscript{2} \\
    \textbf{Duochao Shi\textsuperscript{1}} \quad
    \textbf{Akide Liu\textsuperscript{2}} \quad
    \textbf{Bohan Zhuang\textsuperscript{1}}\\[1em]
    \textsuperscript{1}ZIP Lab, Zhejiang University \quad
    \textsuperscript{2}Monash University
}

\begin{document}

\footnotetext[1]{Corresponding authors: wangweijie@zju.edu.cn, donny.chen@outlook.sg.} 
\footnotetext[2]
{This work was conducted while D. Y. Chen was affiliated with Monash University.} 

\maketitle

\begin{abstract}

Feed-forward 3D Gaussian Splatting (3DGS) models have recently emerged as a promising solution for novel view synthesis, enabling one-pass inference without the need for per-scene 3DGS optimization. However, their scalability is fundamentally constrained by the limited capacity of their models, leading to degraded performance or excessive memory consumption as the number of input views increases. In this work, we analyze feed-forward 3DGS frameworks through the lens of the Information Bottleneck principle and introduce \method, a lightweight \emph{architecture-agnostic module} that enables efficient compression of multi-view inputs into a compact latent state $Z$ that retains essential scene information while discarding redundancy. Concretely, \method enables existing feed-forward 3DGS models to scale to over 100 input views at 480P resolution on an 80GB GPU, by partitioning the views into anchor and support sets and using cross attention to compress the information from the support views into anchor views, forming the compressed latent state $Z$. We show that integrating \method into several state-of-the-art feed-forward 3DGS models consistently improves performance under moderate input views and enhances robustness under dense view settings on two large-scale benchmarks DL3DV-10K and RealEstate10K. The video results, code and trained models are available on our project page: \url{https://lhmd.top/zpressor}.

\end{abstract}

\section{Introduction}

Novel view synthesis (NVS) has played an important role in many everyday applications and is expected to become even more crucial in the future as a foundational technique for augmented reality (AR) and virtual reality (VR). It has also received growing attention in the research community with the introduction of 3D Gaussian Splatting (3DGS)~\cite{kerbl20233d} and a series of subsequent developments~\cite{huang20242d,moenne20243d,mai2024ever,condor2025don,govindarajan2025radiant}. Although 3DGS achieves real time rendering and high visual quality, its reliance on slow per-scene tuning significantly limits its practical use in real world scenarios.

\begin{figure}
    \centering
    \includegraphics[width=\linewidth]{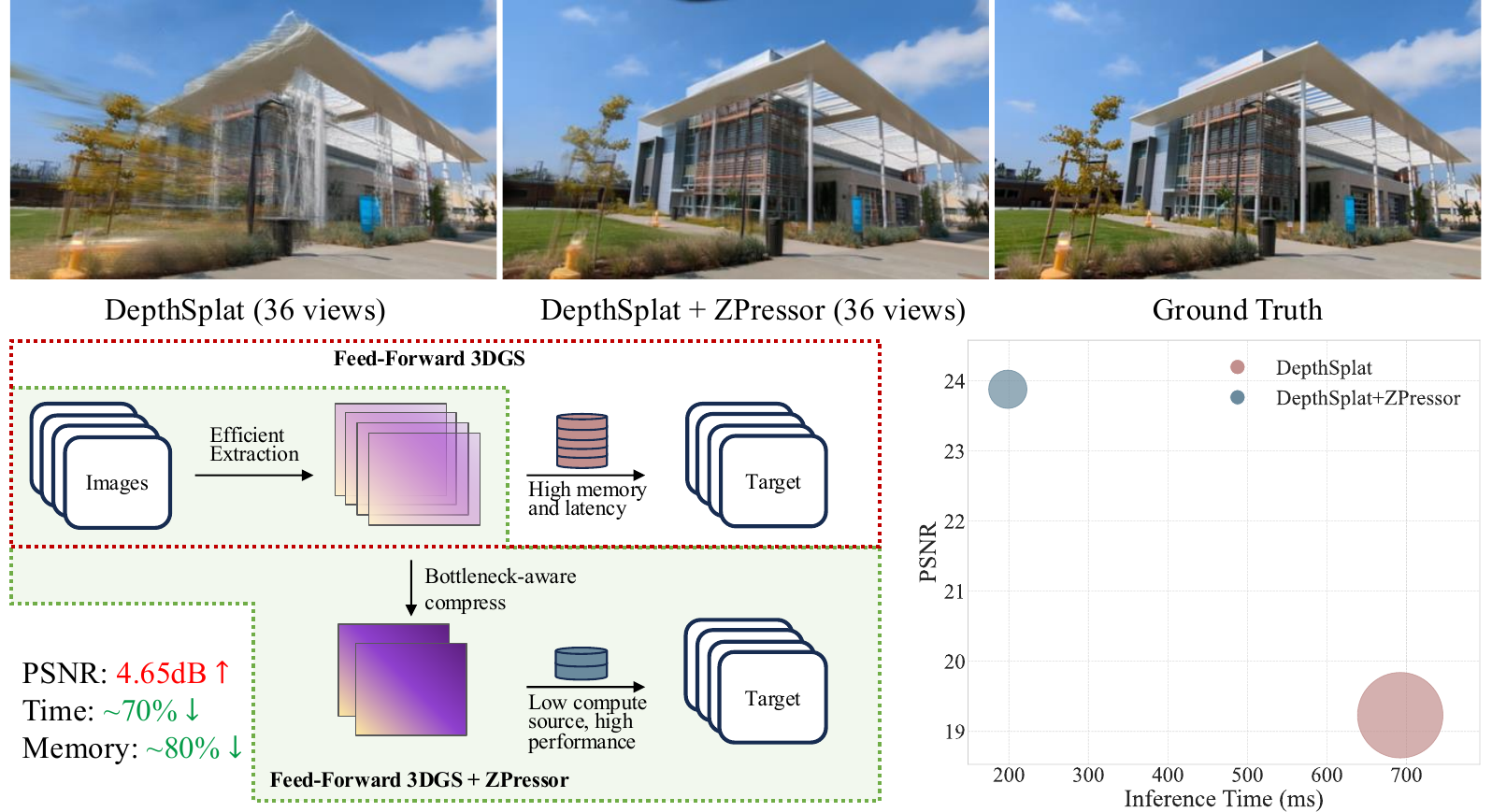}
    \caption{ We visualize the result of DepthSplat~\cite{xu2024depthsplat} with 36 view input for novel view synthesis and after adding ZPressor.  We report PSNR, inference time, and memory usage before and after simply adding the \method, where the radius of the bubble corresponds to memory.}
    \label{fig:teaser}
    \vspace{-0.2cm}
\end{figure}

To address this limitation, feed-forward 3DGS~\cite{charatan2024pixelsplat,chen2024mvsplat} has been introduced to improve the usability of 3DGS. Unlike conventional 3DGS approaches that rely on slow per-scene backward optimization, feed-forward 3DGS introduces an ``encoder'' to extract scene dependent features from input images, allowing the model to benefit from large scale training and predict 3DGS in a single forward pass. Despite notable progress~\cite{wewer2024latentsplat,wang2024freesplat,zhang2024gaussian,xu2024depthsplat,ziwen2024long}, these methods remain constrained to a small number of input views, limiting their ability to fully utilize datasets with dense multiple input views~\cite{reizenstein2021common,yeshwanth2023scannet++,yu2023mvimgnet,ling2024dl3dv}. For example, our experiments show that the state-of-the-art model DepthSplat~\cite{xu2024depthsplat} suffers a significant performance drop and increased computational cost as input views become denser
(see \cref{tab:multi view results dl3dv} and Fig.~\ref{fig:teaser}), . While better engineering might alleviate this memory issue to some extent, it cannot address the huge performance degradation. Upon examining the architecture of several representative feed-forward 3DGS models~\cite{charatan2024pixelsplat,chen2024mvsplat,xu2024depthsplat}, we identify the \emph{limited capacity of the whole network from image to 3DGS} as the root cause.By design, it struggles to scale to denser inputs due to representation overload from excessive feature tokens and the resulting high computational cost.

Rather than introducing yet another ad-hoc encoder design, this work revisits the feed-forward 3DGS framework with inspiration from the Information Bottleneck (IB)~\cite{tishby2000information} principle. 
IB offers a theoretical foundation for learning compact representations that preserve only task-relevant information. In the context of NVS with dense input views, we hypothesize that a latent representation can be learned to capture essential scene information while discarding redundant details in dense multi-view inputs. 
By encouraging the formation of such a compressed yet informative representation, we aim to improve the scalability of feed-forward 3DGS models (detailed in~\cref{sec:info_analysis}). Building on this insight, we propose a lightweight module, termed \method, designed to be seamlessly integrated into the encoder of existing feed-forward 3DGS models to enhance their scalability. Unlike typical efforts that might rely on engineering optimizations such as memory-efficient attention~\cite{dao2023flashattention} or activation checkpointing~\cite{chen2016training}, our approach adopts a principled perspective grounded in representation learning, aiming to address core architectural limitations under dense input views settings.

To put this idea into practice, \method implements the IB principle by explicitly compressing input view information, as illustrated in Fig.~\ref{fig:pipeline}. Specifically, we divide the input views into two groups: anchor views and support views. Anchor views serve as the compression states, while information from support views is compressed into them. To ensure the compressed representation retains sufficient scene information, we select anchor views using farthest point sampling to maximize coverage with restricted views. The remaining views are assigned to their nearest anchor based on camera distance,  and their features are fused into the anchors through a stack of customized cross-attention blocks. In essence, \method takes multiple input views' features and their corresponding camera poses as input and produces a compact latent representation that preserves scene information. This design is architecture-agnostic and thus can be integrated into various feed-forward 3DGS models.

To validate the effectiveness of \method, we integrate \method into several state-of-the-art feed forward 3DGS models, including pixelSplat~\cite{charatan2024pixelsplat}, MVSplat~\cite{chen2024mvsplat}, and DepthSplat~\cite{xu2024depthsplat}, and conduct extensive experiments on large-scale benchmarks such as DL3DV-10K~\cite{ling2024dl3dv} and RealEstate-10K~\cite{zhou2018stereo}.
Results show that integrating \method consistently boosts the performance of baseline models under a moderate number of input views (\eg, 12 views), and helps them maintain reasonable accuracy and computational cost even with very dense inputs (\eg, 36 views, as shown in \cref{fig:teaser}), where the original models typically degrade dramatically or run out of memory. Our contributions are threefold:

\begin{itemize}[leftmargin=*]
    \item We provide a fundamental analysis of why existing feed-forward 3DGS models struggle with dense input views, through the lens of the Information Bottleneck principle.
    \item Inspired by IB, we propose \method, an architecture-agnostic module that can be integrated into the encoder of existing feed-forward 3DGS models to compress input view information.
    \item Extensive experiments on several large-scale benchmarks show that \method consistently improves the performance of baseline models with a moderate number of input views, and further enhances robustness under dense input settings, where existing models typically degrade significantly.
\end{itemize}

\section{Related Work}

\subsection{Information bottleneck and its applications}
The challenge of managing and processing vast quantities of information is a central theme in the development of large-scale machine learning models, particularly in the visual domain~\cite{dosovitskiy2020vit, darcet2023vitneedreg, caron2021emerging, oquab2023dinov2}. 
The Information Bottleneck principle~\cite{tishby2000information} formalizes the problem of extracting a compressed representation $Z$ from the input $X$, such that $Z$ is maximally informative about the target $Y$.
The IB principle was subsequently extended to the domain of deep learning~\cite{mahabadi2021variational, pmlr_v80_dai18d, Kim_2020_CVPR_Workshops, bai2025rethinking}, VIB~\cite{alemi2016deep} providing a tractable lower bound on the IB objective, bridging the gap between the theoretical IB principle and practical deep learning applications.
A series of works have applied the IB principle to multi-view inputs~\cite{federici2020learningrobustrepresentationsmultiview, zhang2024discoveringcommoninformationmultiview, xie2025gaussmigaussiansplattingshannon}, extract information that is common or shared across multiple views while discarding view-specific or redundant information.
The drive towards efficient 3D scene reconstruction, especially within the context of 3DGS~\cite{kerbl20233d}, has also seen the adoption of information-theoretic ideas. 
StreamGS~\cite{li2025streamgsonlinegeneralizablegaussian} tackle redundancy in image streams by merging superfluous Gaussians through cross-frame feature aggregation, while other works~\cite{liu2024compgsefficient3dscene,fan2024lightgaussianunbounded3dgaussian} focus on compressing the learned 3D Gaussians. While demonstrating the effectiveness and necessity of compression in multi-view data and 3D reconstruction, none of the existing works have explored it in the context of feed-forward 3DGS. Our work aims to bridge this gap by introducing the lens of IB for information compression in this area.

\subsection{Optimization-based NeRF and 3DGS}
Traditional novel view synthesis (NVS) methods primarily rely on image blending techniques~\cite{chen2023view, seitz1996view}. More recently, neural network-based approaches~\cite{lombardi2019neural,sitzmann2019scene,sitzmann2021light,mildenhall2020nerf} have advanced NVS by integrating it with deep learning models. In particular, NeRF~\cite{mildenhall2020nerf} employs an MLP to map 3D spatial locations and viewing directions to radiance color and volume density.
Numerous works~\cite{wang2022r2l,sun2021direct,kangle2021dsnerf,muller2022instant,chen2022tensorf,cao2024lightning} have sought to improve NeRF's efficiency and reconstruction quality. However, its reliance on volume rendering~\cite{kajiya1984ray} hinders rendering speed, limiting its practicality in real-world applications.
Recently, 3D Gaussian Splatting (3DGS)\cite{kerbl20233d} and its variants\cite{huang20242d,fan2024instantsplat,lu2024scaffold,girish2024eagles,fan2024lightgaussianunbounded3dgaussian} have emerged as efficient solutions for large-scale scene reconstruction and synthesis, offering explicit representations and fast rasterization-based rendering that outperform NeRF's slower volumetric approach.
Nonetheless, its requirement for slow per-scene optimization still poses challenges for deployment in downstream tasks.

\subsection{Feed-Forward NeRF and 3DGS}
To address the limitation of slow per-scene optimization, PixelNeRF~\cite{yu2021pixelnerfneuralradiancefields} pioneered the feed-forward NeRF (\aka generalizable NeRF) by introducing an additional network that directly encodes input views into a NeRF representation. This allows the model to benefit from large-scale training and predict a scene representation in a single forward pass. This direction has since seen significant advancements~\cite{wang2021ibrnetlearningmultiviewimagebased, chen2021mvsnerffastgeneralizableradiance,johari2022geonerfgeneralizingnerfgeometry,chen2023explicit,xu2024murfmultibaselineradiancefields}, offering a promising path toward practical NeRF deployment.
This paradigm has recently been extended to real-time rendering with 3DGS~\cite{kerbl20233d} replacing NeRF~\cite{mildenhall2020nerf}. Among them, pixelSplat~\cite{charatan2024pixelsplat} pioneered the feed-forward 3DGS approach by combining epipolar transformers with depth prediction to predict 3D Gaussians from two input views. MVSplat~\cite{chen2024mvsplat} proposed an efficient cost-volume-based fusion strategy to improve multi-view reconstruction, while DepthSplat~\cite{xu2024depthsplat} leveraged monocular depth features to better recover fine 3D structures from sparse inputs.
Despite the growing number of feed-forward 3DGS models~\cite{fei2024pixelgaussian,ye2024no,min2024epipolarfree,kang2025selfsplat, chen2024mvsplat360}, most follow a pixel-aligned design, where the number of 3D Gaussians scales linearly with the number of input views. This leads to significant memory and computational overhead as input views increase. While works like FreeSplat~\cite{wang2024freesplat} and GGN~\cite{zhang2024gaussian} attempt to reduce the number of Gaussians by merging them via cross-view projection checking, they lack a principled framework.
In contrast, our work provides a theoretical perspective on the representation overload problem in feed-forward 3DGS by introducing the IB~\cite{tishby2000information} principle. And we propose an architecture-agnostic module \method that can be seamlessly integrated into existing feed-forward 3DGS models to improve performance under dense input view settings.

\section{Methodology}
~\label{sec:method}
\vspace{-0.6cm}
\subsection{Overview of \method}
\begin{figure}
    \centering
    \includegraphics[width=\linewidth]{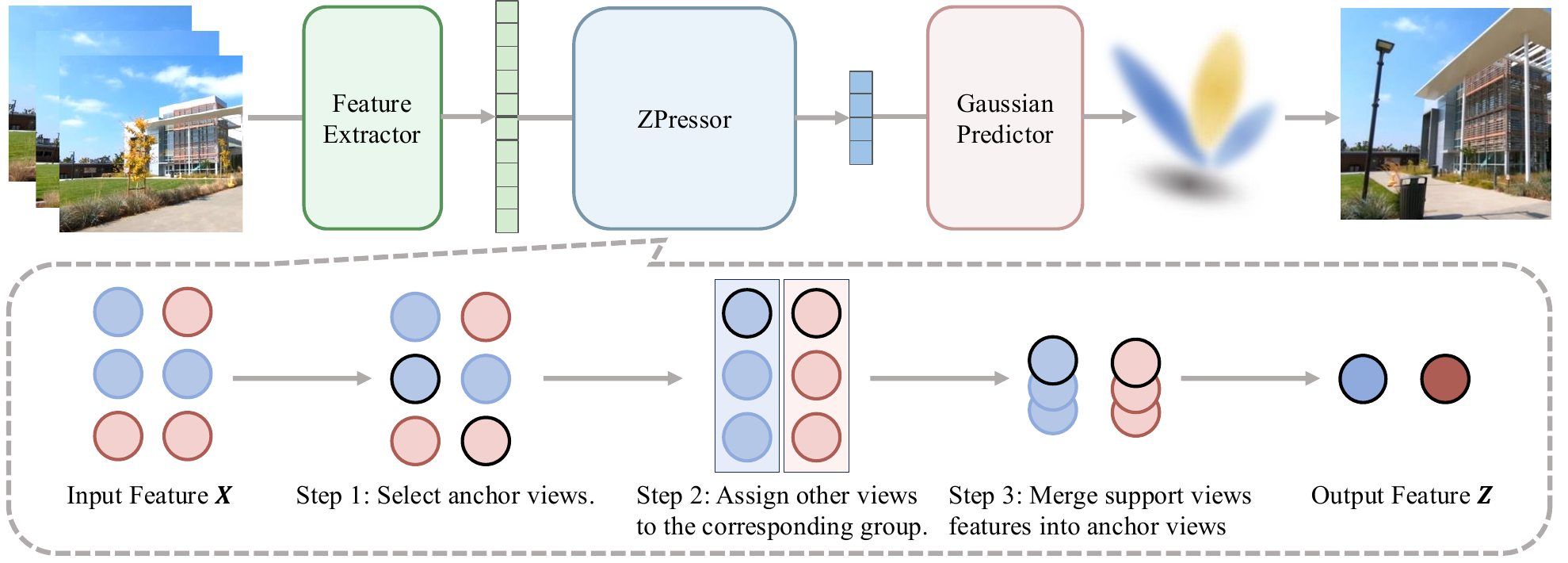}
    \caption{\textbf{Overview of \method for Feed-Forward 3DGS.} Our proposed \method is an architecture-agnostic module designed for feed-forward 3DGS frameworks. It addresses the challenge of processing dense input views by strategically grouping input view features $\mX$ based on selected anchor views, then features within each respective group are compressed as $\mZ$.}
    \label{fig:pipeline}
\end{figure}

Our \method is an architecture-agnostic module for compressing the multi-view inputs of feed-forward 3DGS, as illustrated in \cref{fig:pipeline}. Formally, given $K$ input views $\mathcal{V}=\{{\mathbf V}_{i}\}_{i=1}^K$ where ${\mathbf V}_i \in \mathbb{R}^{H \times W \times 3}$ and their corresponding camera poses $\mathcal{P}=\{{\mathbf P}_{i}\}_{i=1}^K$, our \method takes the extracted features from each view as input: 
\begin{align}
    \label{eq:feature_extraction}
    \mathcal{X}=\{\mathbf{F}_i\}_{i=1}^K=\Phi_\mathrm{image}(\mathcal{V}, \mathcal{P}), \quad \mathbf{F}_i
    \in\mathbb{R}^{\frac{H}{p}\times \frac{W}{p}\times C}
\end{align}
where $\Phi_\mathrm{image}$ is a pretrained image encoder.
Then our \method adaptively compresses these heavy multi-view features into compact ones $\mZ = \text{\method}(\mathcal{\mX})$.
Subsequently, we directly unprojects these compact latent representations into 3D space using the camera poses $\mathcal{P}$ and a pixel-aligned Gaussian prediction network $\Psi_\mathrm{pred}$ is employed to estimate the Gaussian parameters:

\begin{align}
    \mathcal{Y} = \left\{(\mathbf{\mu}_{i}, \mathbf{\Sigma}_{i}, \mathbf{\alpha}_{i}, \mathbf{c}_{i}\right)\}_{i=1}^{H \times W \times K} = \Psi_\mathrm{pred}(\mZ, \mathcal{P}). 
\end{align}
The Gaussian parameters $\mY$ include mean $\mathbf{\mu}$, opacity $\mathbf{\alpha}$, covariance matrix $\mathbf{\Sigma}$, and color $\mathbf{c}$, while this pixel-aligned prediction results in a linear increase in Gaussian primitives with more input views, constraining the model's input capacity.

\subsection{Information Analysis of Feed-Forward 3DGS} \label{sec:info_analysis}

Existing feed-forward 3D Gaussian Splatting networks suffer from a dramatic performance drop and 
an exponential
increase in computational cost when the amount of input view information grows (see ~\cref{tab:multi view results dl3dv}), primarily due to information redundancy and the lack of adaptive information compression mechanisms. Specifically, the total information of the scene is represented by the joint entropy $H(\mathbf{F}_1,\mathbf{F}_2,...,\mathbf{F}_K)$, which is not merely the sum of the individual entropies of the features from all views, \ie, $\sum_{i=1}^K H(\mathbf{F}_i)$.  Therefore, there is a significant amount of redundant information in the features, and it is crucial to remove the irrelevant information after feature extraction while preserving its predictive power, which allows for the efficient use of information
from the input views.

A priciple way for modeling this is the Information Bottleneck (IB)~\cite{tishby2000information}, which minimizes the IB scores as:

\begin{align}
    \min_{{\mZ}} IB =\underbrace{\beta\, I(\mX,~\mZ)}_{\text{Compression Score}}-\underbrace{I(\mZ,~\mY)}_{\text{Prediction Score}}, 
    \label{eq:ib}
\end{align} 

where $\beta \geq 0$ controls the balance between compression and prediction, and $I(\cdot, \cdot)$ is the mutual information. 
The \emph{Compression Score} component, $\beta I(\mX,\mZ)$, encourages $\mZ$ to be a concise representation of the input $\mX$. Minimizing $I(\mX,\mZ)$ means reducing the amount of information $\mZ$ carries about $\mX$, which leads to better compression and enhanced efficiency. The \emph{Prediction Score} component, $I(\mZ,\mY)$, measures the predictive power of the latent feature $\mZ$ with respect to the target variable $\mY$. Maximizing this term ensures that $\mZ$ retains sufficient task-relevant information about $\mY$, which is vital for maintaining or improving prediction accuracy. 

As shown in \cref{fig:pipeline}, the \emph{Prediction Score} is typically modelled by the Gaussian predictor. In the next section, we introduce \method, a novel module applied concatenated with the image encoder to model the \emph{Compression Score} through three consecutive designs, yielding a compact presentation.

\subsection{Information Compression Module: \method}

To ensure architecture-agnostic integration, compression is performed along the view dimension, rather than being entangled with the design of a specific model. 
Concretely, given the $K$ encoded views $\mX$,
we first divide them into $N$ \emph{anchor views} $\mathcal{X}_\mathrm{anchor} = \{\mathbf{F}_{a_i}\}_{i=1}^N$ and $M$ \emph{support views} $\mathcal{X}_\mathrm{support} = \{\mathbf{F}_{s_j}\}_{j=1}^M$, where $a_i, s_j \in \{1, \ldots, K\}$,  $\mathcal{X}_\mathrm{anchor} \cap \mathcal{X}_\mathrm{support} = \emptyset$, and $\mathcal{X}_\mathrm{anchor} \cup \mathcal{X}_\mathrm{support} = \mathcal{X}$.
Here, $\mathcal{X}_\mathrm{anchor}$ are expected to capture the essential information of the scene, while $\mathcal{X}_\mathrm{support}$ contain supporting context but may include redundancy. Compression is then achieved by \emph{fusing} the features of support views into their \emph{corresponding} anchor views. This raises three design questions: \textbf{1)} how to select the anchors $\mathcal{X}_\mathrm{anchor}$, \textbf{2)} how to assign $\mathcal{X}_\mathrm{support}$ to specific anchors, and \textbf{3)} how to fuse the information from $\mathcal{X}_\mathrm{support}$ into their designated $\mathcal{X}_\mathrm{anchor}$.

\noindent\textbf{Anchor view selection} (\cref{fig:pipeline} Step~1).
Given a set of camera positions  $\mathcal{T} = \{\mathbf{T}_{i}\}_{i=1}^K \in \mathbb{R}^{K \times 3}$ calculated from camera parameters $\mathcal{P}$, where $f \colon \mathbf{T} \to \mathbf{F}$ such that $f$ is a bijective mapping, 
we first add a random view to the anchor view list $\mathcal{S} = \{\mathbf{T}_{a_1}\}$, where $\mathbf{T}_{a_1} \sim \text{Uniform}(\mathcal{T})$.

Subsequent anchor views are iteratively selected as the view with the greatest distance to the current anchor view set $\mathcal{S}$:

\begin{align}
\mathbf{T}_{a_{i+1}} = \arg\max_{\mathbf{T}_j \in \mathcal{T} \setminus \mathcal{S}} \left( \min_{\mathbf{T}_k \in \mathcal{S}} d(\mathbf{T}_j, \mathbf{T}_k) \right),
\end{align}

where $d(\cdot, \cdot)$ denotes the Euclidean distance. This procedure is repeated until $N$ anchors are selected, resulting in a diverse and representative set $\mathcal{X}_\mathrm{anchor}$.

\noindent\textbf{Support-to-anchor assignment} (\cref{fig:pipeline} Step~2).
Once anchors are selected, each support view is assigned to its nearest anchor based on camera position. This ensures that support views, which capture complementary scene details, are grouped with the most spatially relevant anchor views, thereby ensuring the effectiveness of information fusion. Formally, the cluster assignment to the $i$-th anchor view can be denoted as: 
\begin{align}
\mathcal{C}_i = \{f(\mathbf{T}) \in \mathcal{X}_\mathrm{support} \mid \|\mathbf{T} - \mathbf{T}_{a_i}\| \leq \|\mathbf{T} - \mathbf{T}_{a_j}\|, \forall j \neq i \}
\label{eq:assign}
\end{align}

\noindent\textbf{Views information fusion} (\cref{fig:pipeline} Step~3).
Upon obtaining the anchor-support view-based clusters, we aim to fuse the information within each cluster. The fusion mechanism should satisfy the following properties: \textbf{1)} use the anchor views as the base, while effectively integrating information from the support views to enhance them, and \textbf{2)} capture the similarity between the two sets of views, maintaining compactness while avoiding redundancy.

Building on the above analysis, the fusion within each cluster can be effectively achieved via cross-attention:

\begin{align}
   \mZ = \text{Attention}(Q, K, V),
    \quad Q \leftarrow \mathcal{X}_\mathrm{anchor}, \quad K, V \leftarrow \mathcal{X}_\mathrm{support},
\end{align}

where the features $\mathcal{X}_\mathrm{anchor}$ extracted from the anchor views are used as queries, while the features $\mathcal{X}_\mathrm{support}$ from the support views provide keys and values.
In this way, information from the support views is effectively integrated into the anchor views, satisfying the first property. Additionally, it ensures gradient flow from the prediction to both anchor and support views, enabling the model to capture correlations between the two sets of views, thus satisfying the second property. Moreover, since most existing feed-forward 3DGS encoders adopt Transformer-based architectures, the attention module can be readily integrated into these frameworks.

\noindent\textbf{Training.} Upon obtaining the compressed information $\mZ$, %
we can apply the IB principle to regularize the information flow. Recall that, as in \cref{eq:ib}, our current goal should be to minimize the \emph{Compression Score} $I({\mathcal{X}_\mathrm{anchor}, \mathcal{X}_\mathrm{support}}; \mZ)$ and maximize \emph{Prediction Score} $I(\mZ; \mY)$. For the \emph{Compression Score}, we apply a constraint on its complexity, \ie, setting the number of anchor views $N$ to a value acceptable for training. For the \emph{Prediction Score}, we have
\begin{align}
    I(\mZ; \mY) = H(\mY) - H(\mY \mid \mZ), 
\end{align}
where $H(\mY)$ 
is a constant representing the 3D Gaussians that model the underlying 3D scene. Hence, maximizing $I(\mZ; \mY)$ is equivalent to minimizing $H(\mY \mid \mZ)$, which essentially encourages the predicted 3D Gaussians $\mY$ to resemble the original scene.

Then we incorporate the IB principle into feed-forward 3DGS training, using the efficient estimate of \cref{eq:ib} following \cite{alemi2016deep, mahabadi2021variational}. Formally, the training objective is defined as:

\begin{align}
    \mathcal{L} = \underset{\mathcal{Z} \sim p_\theta(\mathcal{Z} \mid \mathcal{X})}{\mathbb{E}} \bigl[ -\log q_\phi(\mathcal{Y} \mid \mathcal{Z}) \bigr]  + \beta \, \underset{\mathcal{X}}{\mathbb{E}} \bigl[ \text{KL} \bigl[ p_\theta(\mathcal{Z} \mid \mathcal{X}), \, r(\mathcal{Z}) \bigr] \bigr]
\end{align}

where $\phi$ denotes the parameters of 3D Gaussians prediction network and $-\log q_\phi(\mathcal{Y} \mid \mathcal{Z})$ can be modeled by the rendering loss (such as the MSE and LPIPS loss), $\theta$ denotes the parameters of the network preceding $\mathcal{Z}$ and $p_\theta(\mathcal{Z} \mid \mathcal{X})$ is the posterior probability estimate of $\mathcal{Z}$, $r(\mathcal{Z}) \sim \mathcal{N}(\mathcal{Z} \mid \mu_0, \Sigma_0)$ is the Gaussian prior of $\mathcal{Z}$. 

In our implementation, we append an additional self-attention layer to further enhance information flow within each cluster. In addition, we stack several blocks (each containing both cross- and self-attention) to further improve the effectiveness of the IB principle. These two general-purpose engineering techniques help boost performance, as shown in \cref{tab:ablation}. 

\section{Experiments}
\label{sec:experiments}
\subsection{Experimental Settings}
\noindent\textbf{Datasets.}
We validate the effectiveness of our \method on the NVS task, following existing works~\cite{charatan2024pixelsplat,chen2024mvsplat,xu2024depthsplat}, and conduct experiments on two large-scale datasets: DL3DV-10K (DL3DV)~\cite{ling2024dl3dv} and RealEstate10K (RE10K)~\cite{zhou2018stereo}.
DL3DV is a challenging large-scale dataset that contains 51.3 million frames from 10,510 real scenes. We used 140 benchmark scenes for testing and the remaining 9896 scenes for training, with filtering applied to ensure that there is strictly no overlap between the training and test sets.
RE10K offers a large-scale collection of indoor home tour clips, comprising 10 million frames from around 80,000 video clips sourced from public YouTube videos. It is split into 67,477 training and 7,289 testing scenes.
Both datasets feature real-world captured scenes, with camera intrinsics and extrinsics reconstructed using COLMAP~\cite{schoenberger2016sfm,schoenberger2016mvs}.

\noindent \textbf{Baselines and metrics.}
To evaluate the effectiveness and flexibility of our proposed \method, we integrate it as a module into three representative baselines, including DepthSplat~\cite{xu2024depthsplat}, MVSplat~\cite{chen2024mvsplat}, and pixelSplat~\cite{charatan2024pixelsplat}. For fair comparisons, we insert \method into the official implementations of each baseline and strictly follow the experimental settings described in their respective papers. Specifically, we compare with DepthSplat on DL3DV, following its training strategy by first pre-training on RE10K and then fine-tuning on DL3DV. Comparisons with MVSplat and pixelSplat are conducted on RE10K. Following these baselines, we report quantitative results using PSNR, SSIM~\cite{wang2004image}, and LPIPS~\cite{zhang2018unreasonable}. As our module focuses on information compression, we additionally report model efficiency in terms of runtime and memory consumption.

\begin{table}[htbp]
  \centering
  \caption{\textbf{Quantitative comparisons on DL3DV~\cite{ling2024dl3dv}.} We evaluate both DepthSplat~\cite{xu2024depthsplat} and DepthSplat with \method with 12, 16, 24, 36 input views and test on eight target novel views.}
  \vspace{0.4cm}
    \scalebox{0.9}{
  \setlength{\tabcolsep}{1.5mm}{\begin{tabular}{llccc}
    \toprule
    Views & Methods & PSNR$\uparrow$ & SSIM$\uparrow$ & LPIPS$\downarrow$ \\
    \midrule
    \multirow{2}{*}{36 views}& DepthSplat & 19.23 & 0.666 & 0.286 \\
    & $\text{DepthSplat + ZPressor}$ & \phantom{\bf \tiny +0.01}\hfill \textbf{23.88}{\raggedleft  \color{soft_red}       \bf \tiny +4.65} & \phantom{\bf \tiny +0.001}\hfill \textbf{0.815}{\raggedleft  \color{soft_red}       \bf \tiny +0.149} & \phantom{\bf \tiny +0.001}\hfill \textbf{0.150}{\raggedleft  \color{soft_red}       \bf \tiny -0.136} \\
    \midrule
    \multirow{2}{*}{24 views}& DepthSplat & 20.38 & 0.711 & 0.253 \\
    & $\text{DepthSplat + ZPressor}$ & \phantom{\bf \tiny +0.01}\hfill \textbf{24.26}{\raggedleft  \color{soft_red}       \bf \tiny +3.88} & \phantom{\bf \tiny +0.001}\hfill \textbf{0.820}{\raggedleft  \color{soft_red}       \bf \tiny +0.109} & \phantom{\bf \tiny +0.001}\hfill \textbf{0.147}{\raggedleft  \color{soft_red}       \bf \tiny -0.106} \\
    \midrule
    \multirow{2}{*}{16 views}& DepthSplat & 22.07 & 0.773 & 0.195 \\
    & $\text{DepthSplat + ZPressor}$ & \phantom{\bf \tiny +0.01}\hfill \textbf{24.25}{\raggedleft  \color{soft_red}       \bf \tiny +2.18} & \phantom{\bf \tiny +0.001}\hfill \textbf{0.819}{\raggedleft  \color{soft_red}       \bf \tiny +0.046} & \phantom{\bf \tiny +0.001}\hfill \textbf{0.147}{\raggedleft  \color{soft_red}       \bf \tiny -0.047} \\
    \midrule
    \multirow{2}{*}{12 views}& DepthSplat & 23.32 & 0.807 & 0.162 \\
    & DepthSplat + \method & \phantom{\bf \tiny +0.01}\hfill \textbf{24.30}\scriptsize{\raggedleft  \color{soft_red}       \bf \tiny +0.97} & \phantom{\bf \tiny +0.001}\hfill \textbf{0.821}{\raggedleft  \color{soft_red}       \bf \tiny +0.014} & \phantom{\bf \tiny +0.001}\hfill \textbf{0.146}{\raggedleft  \color{soft_red}       \bf \tiny -0.017} \\
    \bottomrule
  \end{tabular}}}
  \label{tab:multi view results dl3dv}
\end{table}

\begin{table}[htbp]
  \centering
  \caption{\textbf{Quantitative comparisons on RE10K~\cite{zhou2018stereo}.} We test pixelSplat~\cite{charatan2024pixelsplat} and MVSplat~\cite{chen2024mvsplat} on eight target views, "OOM" represent that model cannot infer on an 80G GPU.} 
  \vspace{0.4cm}
      \scalebox{0.9}{
  \setlength{\tabcolsep}{1.5mm}{\begin{tabular}{llccc} %
    \toprule
    Views & Methods & PSNR$\uparrow$ & SSIM$\uparrow$ & LPIPS$\downarrow$ \\
    \midrule
    \multirow{4}{*}{36 views}& pixelSplat & OOM & OOM & OOM \\
    & $\text{pixelSplat + ZPressor}$ & \textbf{26.59} & \textbf{0.849} & \textbf{0.225} \\
    & $\text{MVSplat}$ & 24.19 & 0.851 & 0.155 \\
    & $\text{MVSplat + ZPressor}$ & \phantom{\bf \tiny +0.01}\hfill \textbf{27.34}{\raggedleft \color{soft_red} \bf \tiny +3.15} & \phantom{\bf \tiny +0.001}\hfill \textbf{0.893}{\raggedleft \color{soft_red} \bf \tiny +0.042} & \phantom{\bf \tiny +0.001}\hfill \textbf{0.113}{\raggedleft \color{soft_red} \bf \tiny -0.042} \\
    \midrule
    \multirow{4}{*}{24 views} & $\text{pixelSplat}$ & OOM & OOM & OOM \\
    & $\text{pixelSplat + ZPressor}$ & \textbf{26.72} & \textbf{0.851} & \textbf{0.223} \\
    & $\text{MVSplat}$ & 25.00 & 0.871 & 0.137 \\
    & $\text{MVSplat + ZPressor}$ & \phantom{\bf \tiny +0.01}\hfill \textbf{27.49}{\raggedleft \color{soft_red} \bf \tiny +2.49} & \phantom{\bf \tiny +0.001}\hfill \textbf{0.895}{\raggedleft \color{soft_red} \bf \tiny +0.024} & \phantom{\bf \tiny +0.001}\hfill \textbf{0.111}{\raggedleft \color{soft_red} \bf \tiny -0.026} \\
    \midrule %
    \multirow{4}{*}{16 views}& pixelSplat & OOM & OOM & OOM \\
    & $\text{pixelSplat + ZPressor}$ & \textbf{26.81} & \textbf{0.853} & \textbf{0.221} \\
    & $\text{MVSplat}$ & 25.86 & 0.888 & 0.120 \\
    & $\text{MVSplat + ZPressor}$ & \phantom{\bf \tiny +0.01}\hfill \textbf{27.60}{\raggedleft \color{soft_red} \bf \tiny +1.74} & \phantom{\bf \tiny +0.001}\hfill \textbf{0.896}{\raggedleft \color{soft_red} \bf \tiny +0.008} & \phantom{\bf \tiny +0.001}\hfill \textbf{0.110}{\raggedleft \color{soft_red} \bf \tiny -0.010} \\
    \midrule %
    \multirow{4}{*}{8 views}& pixelSplat & 26.19 & 0.852 & \textbf{0.215} \\
    & $\text{pixelSplat + ZPressor}$ & \phantom{\bf \tiny +0.01}\hfill \textbf{26.86}{\raggedleft \color{soft_red} \bf \tiny +0.67} & \phantom{\bf \tiny 0.001}\hfill \textbf{0.854}{\raggedleft \color{soft_red} \bf \tiny +0.002} & \phantom{\bf \tiny +0.001}\hfill 0.219{\raggedleft \color{dark_green} \bf \tiny +0.004} \\
    & $\text{MVSplat}$ & 26.94 & \textbf{0.902} & \textbf{0.107} \\
    & $\text{MVSplat + ZPressor}$ & \phantom{\bf \tiny +0.01}\hfill \textbf{27.72}{\raggedleft \color{soft_red} \bf \tiny +0.78} & \phantom{\bf \tiny 0.001}\hfill 0.897{\raggedleft \color{dark_green} \bf \tiny -0.005} & \phantom{\bf \tiny +0.001}\hfill 0.109{\raggedleft \color{dark_green} \bf \tiny +0.002} \\
    \bottomrule
  \end{tabular}}}
  \label{tab:multi view results re10k}
\end{table}

\noindent \textbf{Implementation details.}
We use the same computing resources to train the baseline and our method.
Due to the memory limit, we use 6 context views for DepthSplat and MVSplat, and 4 context views for pixelSplat. For all of our experiments, we adopted the same learning rate as the baseline, utilized the AdamW optimizer, and trained the models for 100,000 steps on A800 GPUs. 
Following the setting of the baseline, we use the $256\times 256$ input resolution on RE10K,
and $256\times 448$ input resolution on DL3DV.
All training losses match those of the baseline, with no additional data or regularization introduced. 

\subsection{SoTA Comparisons and IB Analysis}

\noindent \textbf{Comparisons with SoTA models.}
We train all models on DL3DV and RE10K using 12 input views with 6 anchor views set to our \method, and evaluate them under varying numbers of input views ranging from 8 to 36.
As shown in \cref{tab:multi view results dl3dv} and \cref{tab:multi view results re10k}, integrating \method into DepthSplat, MVSplat, and pixelSplat consistently improves their performance across all input view settings and evaluation metrics, demonstrating the effectiveness of our approach.

\begin{figure}[t]
  \centering
  \includegraphics[width=1.0\textwidth]{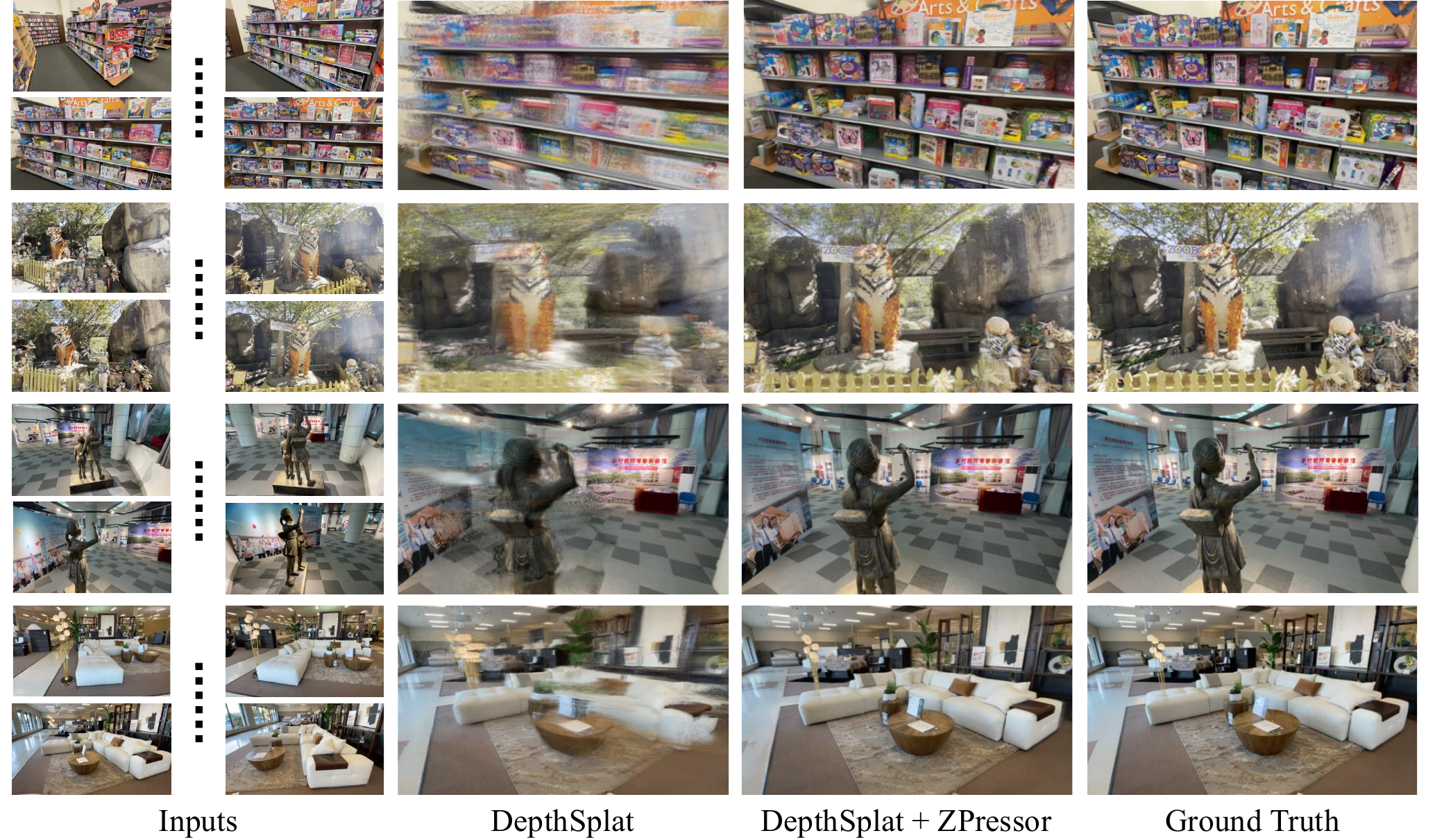}
    \begin{picture}(0,0)
    \put(-21,14){\scriptsize \cite{xu2024depthsplat}}
    \end{picture}
  \caption{\textbf{Qualitative comparison on DL3DV~\cite{ling2024dl3dv} under dense input conditions (36 views).} DepthSplat~\cite{xu2024depthsplat} performs poorly due to redundancy in dense views, \method effectively compresses this information, achieving significantly improved visual results.}
  \label{fig: view_dl3dv} 
\end{figure}
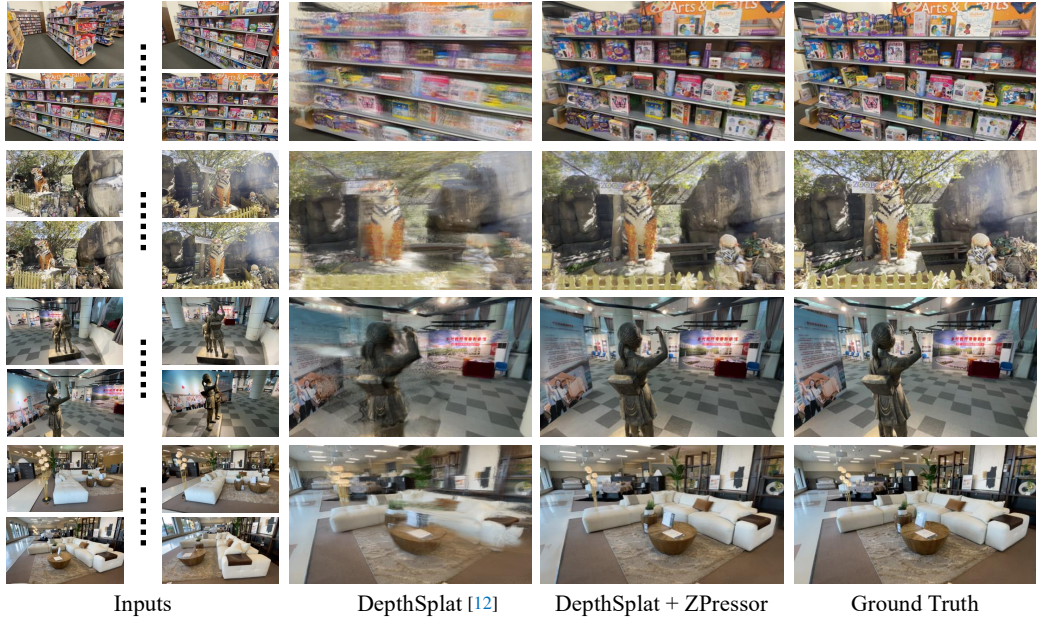

\begin{figure}[t]
  \centering
  \includegraphics[width=1.0\textwidth]{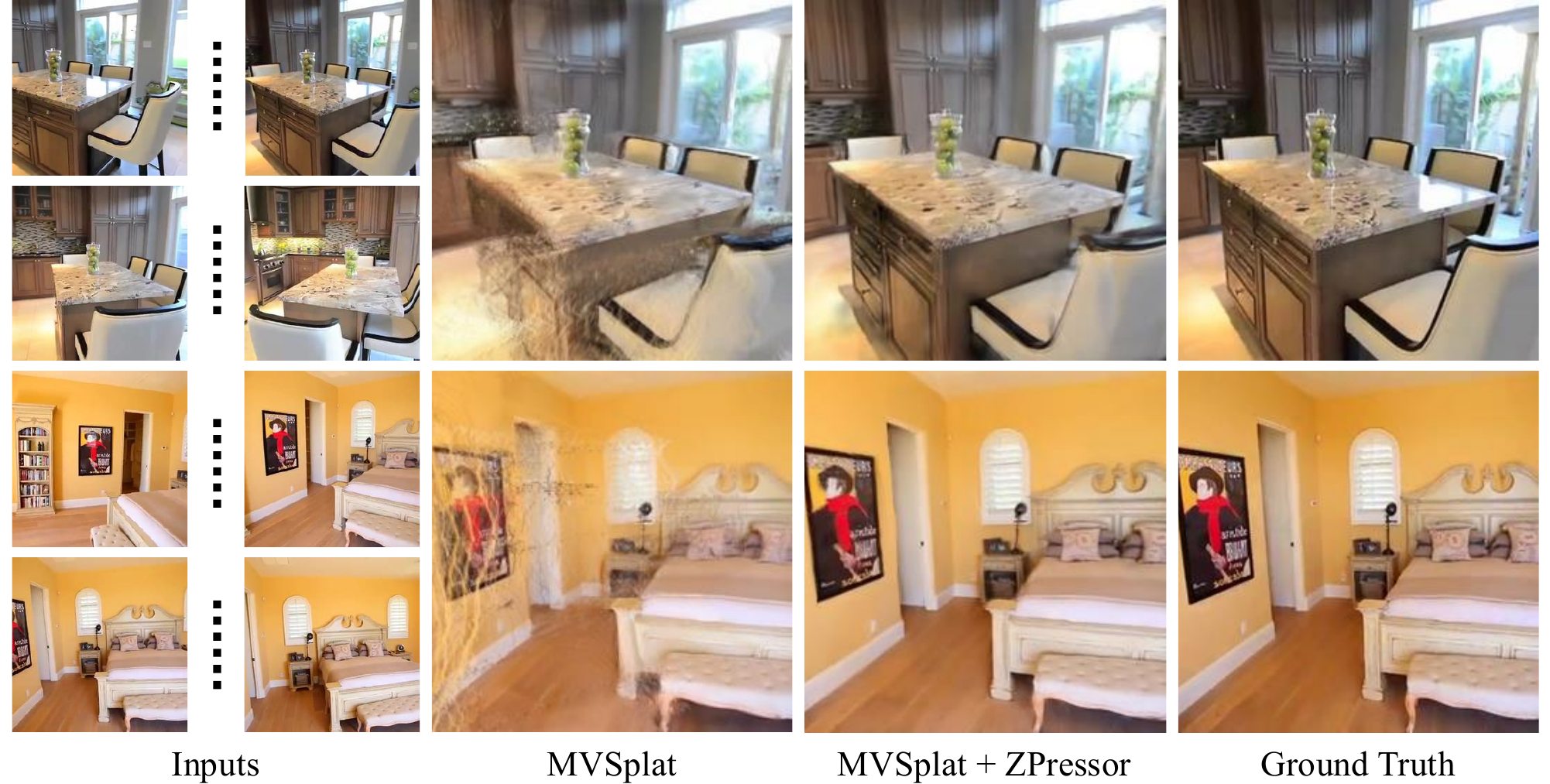}
    \begin{picture}(0,0)
    \put(-22.5,14){\scriptsize \cite{chen2024mvsplat}}
    \end{picture}
  \caption{\textbf{Qualitative comparison on RE10K~\cite{zhou2018stereo} with 36 input views.} MVSplat~\cite{chen2024mvsplat} with \method performs the best in all cases.}
  \label{fig: view_re10k} 
\end{figure}
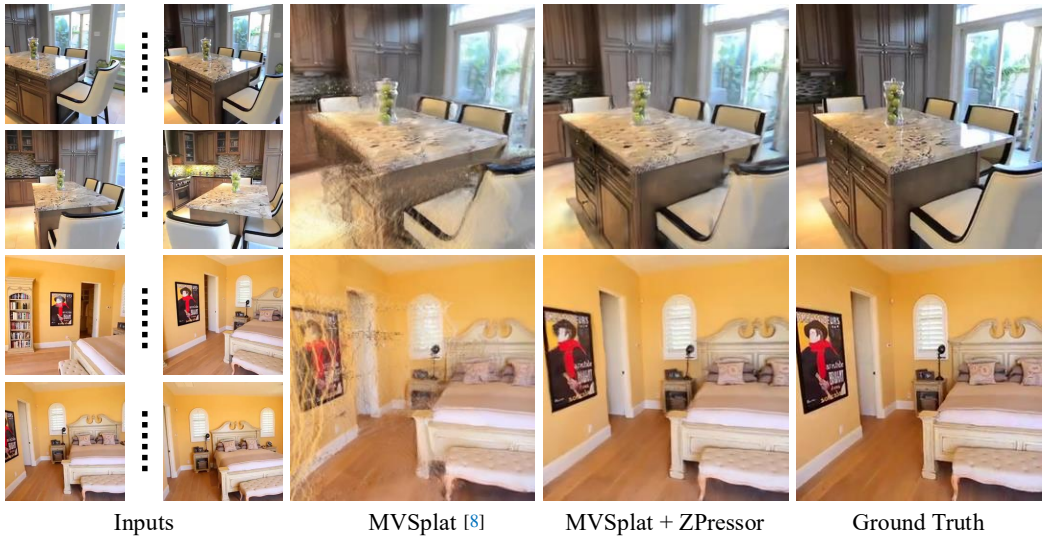

Notably, the performance gain becomes more significant as the number of input views increases. This is because existing feed-forward 3DGS models struggle with dense inputs due to representation overload, leading to performance degradation. In contrast, \method mitigates this issue by compressing the input through redundancy suppression while preserving essential information, improving model robustness, and maintaining strong performance under dense input settings. Moreover, we observe that pixelSplat fails to run with more than 8 input views due to out-of-memory (OOM) caused by the large number of predicted pixel-aligned 3D Gaussians. In contrast, our \method helps merge input information, reducing the number of predicted Gaussians and enabling testing with up to 36 views.
These observations are further supported by the qualitative comparisons shown in \cref{fig: view_dl3dv} and \cref{fig: view_re10k}, where DepthSplat and MVSplat exhibit noticeable artifacts under 36 input views, whereas their \method-augmented versions produce significantly cleaner renderings. 

\begin{figure}[t]
  \centering
  \includegraphics[width=1.0\textwidth]{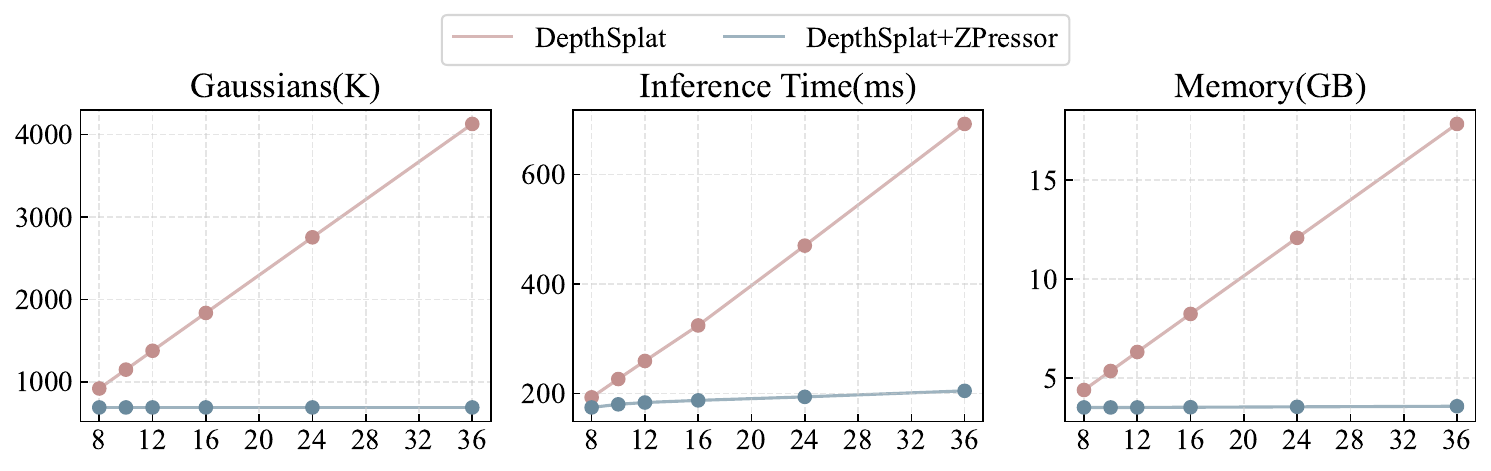}
  \caption{\textbf{Efficiency analysis.} We report the number of Gaussians (K), inference time (ms) and peak memory (GB) of DepthSplat~\cite{xu2024depthsplat} and DepthSplat with \method.} 
  \label{fig: efficiency} 
\end{figure}

\begin{figure}[htbp]
    \centering
    \begin{minipage}[t]{0.48\textwidth}
        \centering
        \includegraphics[width=0.95\linewidth]{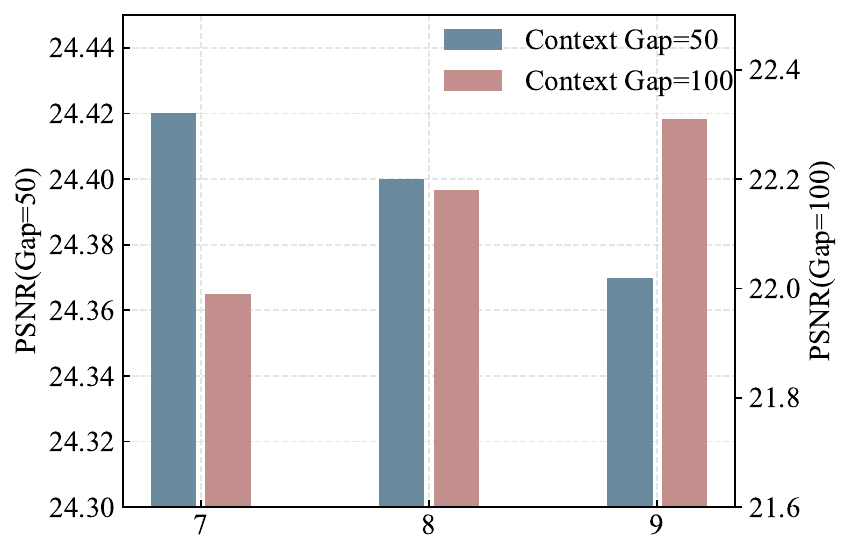}
        \captionof{figure}{\textbf{Analysis of the bottleneck constraint.} We compare the performance of ZPressor in different scale of scene coverage.}
        \label{fig: bottleneck2}
    \end{minipage}
    \hfill
    \begin{minipage}[t]{0.48\textwidth}
        \centering
        \vspace{-4.1cm}
        \captionof{table}{\textbf{Analysis of Information Fusion.}
        ``default'' denotes our setting where support views are fused into anchor views. ``w/o fusion'' removes the fusion step, and ``fuse anchors'' fuses repeated anchor views instead. ``default'' performs best, indicating that \method improves performance by effectively fusing complementary information from the support views.
        }
        \vspace{0.2cm}
        \setlength{\tabcolsep}{2mm}
        \begin{tabular}{cccc}
            \toprule
            Method & PSNR$\uparrow$ & SSIM$\uparrow$ & LPIPS$\downarrow$ \\
            \midrule
            default & \textbf{24.30} & \textbf{0.821} & \textbf{0.146} \\
            fuse anchors & 24.23 & 0.817 & 0.148 \\
            w/o fusion & 23.80 & 0.810 & 0.162 \\
            \bottomrule
        \end{tabular}
        \label{tab:avlation_more_infor}
    \end{minipage}
\end{figure}

\noindent \textbf{Analysis of model efficiency.}
Compressing the input view information not only improves robustness and performance, but also enhances efficiency. To validate this, we compare the model efficiency of DepthSplat with and without \method under 480P resolution, evaluating the number of 3D Gaussians, the test-time inference latency, and the peak memory usage. As shown in \cref{fig: efficiency}, the benefits of integrating \method are clear. In particular, as the number of context views increases, the baseline model's predicted 3D Gaussian numbers, memory usage, and inference time all grow linearly, whereas \method helps maintain stable resource consumption across all aspects.

\noindent \textbf{Analysis of the bottleneck constraint.}
In the context of NVS, the information corresponds to the overall region that the scene covers. Since we evaluated on the static video data DL3DV, longer sequences usually cover larger regions. Therefore, we use the frame distance between input views as a proxy for both scene coverage and information content. 

We can then analysis the effect of bottleneck constraint (the number of anchor views) under scenes with varying information content by adjusting the frame distance between input views.
As shown in \cref{fig: bottleneck2}, we conduct experiments under two settings, Context Gap 50 (CG50, in blue) and Context Gap 100 (CG100, in pink), where context gap refers to the frame distance between input views. 
When the scene's information content is relatively low (e.g., small camera baseline, similar views, as proxied by a small CG like 50), a smaller number of anchor views (e.g., 7) is already sufficient to capture the essential scene information. Adding more anchor views in such a scenario might introduce redundancy or ambiguity, as these additional anchors might not observe genuinely new regions but rather re-observe already covered areas from slightly different perspectives. This could lead to a less compact and potentially noisier latent representation, hence the performance drop. In contrast, for scenes with higher information content (larger CG like 100), more anchor views are beneficial as they help cover more diverse perspectives and capture richer scene details. 
These results highlight the effectiveness of our \method as an instantiation of the IB principle and show that the information bottleneck is critical in balancing compression and information preservation.

\begin{table}[htbp]
  \centering
  \caption{\textbf{Ablation study of our method with DepthSplat~\cite{xu2024depthsplat} on the DL3DV dataset~\cite{ling2024dl3dv}.} Models are evaluated by rendering eight novel views using 12 input views.}
  \vspace{0.4cm}
  \scalebox{0.9}{
  \setlength{\tabcolsep}{2mm}{\begin{tabular}{llccccc}
    \toprule
    Methods & PSNR↑ & SSIM↑ & LPIPS↓ & Time (s) & Peak Memory (GB) \\
    \midrule
    DepthSplat + \method & \textbf{24.30} & \textbf{0.821} & \textbf{0.146} & 0.184 & 3.80 \\
    w/o multi-blocks & 24.18 & 0.817 & 0.149 & \textbf{0.140} & \textbf{3.79} \\
    w/o self-attention & 23.85 & 0.810 & 0.156 & 0.183 & 3.80 \\
    DepthSplat & 23.32 & 0.808 & 0.162 & 0.260 & 6.80 \\
    \bottomrule
  \end{tabular}}}
  \label{tab:ablation}
\end{table}

\noindent \textbf{Analysis of information fusion.}
To further confirm that our \method effectively fuses information from support views into anchor views rather than simply discarding it, we conduct experiments by varying the fusion strategy.
As shown in \cref{tab:avlation_more_infor}, removing the information fusion step (``w/o fusion'') leads to a performance drop, highlighting the importance of fusing support views into anchor views. To ensure that the introduction of support information does result in a performance improvement, we conduct a control experiment by fusing repeated anchor views instead of support views (``fuse anchors''). Since this does not introduce new information, its performance is lower than our default setting, which achieves the best results by fusing complementary information. These comparisons further validate that our design effectively implements the IB principle.

\subsection{Ablation Study}

As mentioned at the end of \cref{sec:method}, the default \method uses several stacked attention blocks, each combining self-attention and cross-attention. This section presents an ablation study to validate our design.

We report results on DL3DV with 12 input views, following the setting in \cref{tab:multi view results dl3dv}, using DepthSplat as the backbone. As shown in \cref{tab:ablation}, removing the stacking design and using only one block (``w/o multi-blocks'') slightly degrades performance, suggesting that stacking improves fusion of information from support to anchor views. Additionally, removing the self-attention (``w/o self attention'') also reduce performance, showing that self-attention complements cross-attention by enhancing internal feature interactions. Overall, all variants of \method outperform the DepthSplat baseline, confirming the existence of an information bottleneck in feed-forward 3DGS models and the effectiveness of our \method in addressing it.

\section{Conclusion}
\label{sec:conclusion}
\vspace{-0.3cm}
We have provided a fundamental analysis of the model capacity limitations in existing feed-forward 3DGS models through the lens of the Information Bottleneck principle. Building on this insight, we have introduced \method, a lightweight, architecture-agnostic module that efficiently compresses multi-view inputs, enabling models to overcome inherent limitations and scale to handle more input views. We have validated our \method by integrating it into several representative feed-forward 3DGS models. Our experiments on several large-scale benchmarks have demonstrated that \method not only consistently improves the performance of existing models under moderate view settings, but also helps them maintain competitive efficiency under denser inputs. We believe that \method significantly enhances the scalability and practicality of feed-forward 3DGS models, opening the door to more effective applications in real-world scenarios.

\noindent\textbf{Limitation and discussion.} Our \method may be less effective in extremely dense view settings. For example, given 1000 input views, \method can only compress them to around 50 views in order to maintain the information compactness as regularized by the IB principle. However, handling 50 views of 3D Gaussians still presents significant computational challenges for typical GPUs. Future work could explore combining \method with 3D Gaussian merging or memory-efficient rendering to extend feed-forward 3DGS to handle extremely dense input views.

\noindent\textbf{Acknowledgements.} We thank Biao Wu for his enlightening discussions and inspiration on network architecture.


\clearpage
\begin{thebibliography}{10}

\bibitem{kerbl20233d}
Bernhard Kerbl, Georgios Kopanas, Thomas Leimk{\"u}hler, and George Drettakis.
\newblock 3d gaussian splatting for real-time radiance field rendering.
\newblock {\em ACM Transactions on Graphics}, 42(4):139--1, 2023.

\bibitem{huang20242d}
Binbin Huang, Zehao Yu, Anpei Chen, Andreas Geiger, and Shenghua Gao.
\newblock 2d gaussian splatting for geometrically accurate radiance fields.
\newblock In {\em ACM SIGGRAPH 2024 conference papers}, pages 1--11, 2024.

\bibitem{moenne20243d}
Nicolas Moenne-Loccoz, Ashkan Mirzaei, Or~Perel, Riccardo de~Lutio, Janick Martinez~Esturo, Gavriel State, Sanja Fidler, Nicholas Sharp, and Zan Gojcic.
\newblock 3d gaussian ray tracing: Fast tracing of particle scenes.
\newblock {\em ACM Transactions on Graphics}, 43(6):1--19, 2024.

\bibitem{mai2024ever}
Alexander Mai, Peter Hedman, George Kopanas, Dor Verbin, David Futschik, Qiangeng Xu, Falko Kuester, Jonathan~T Barron, and Yinda Zhang.
\newblock Ever: Exact volumetric ellipsoid rendering for real-time view synthesis.
\newblock {\em CoRR}, 2024.

\bibitem{condor2025don}
Jorge Condor, Sebastien Speierer, Lukas Bode, Aljaz Bozic, Simon Green, Piotr Didyk, and Adrian Jarabo.
\newblock Don't splat your gaussians: Volumetric ray-traced primitives for modeling and rendering scattering and emissive media.
\newblock {\em ACM Transactions on Graphics}, 2025.

\bibitem{govindarajan2025radiant}
Shrisudhan Govindarajan, Daniel Rebain, Kwang~Moo Yi, and Andrea Tagliasacchi.
\newblock Radiant foam: Real-time differentiable ray tracing.
\newblock {\em arXiv preprint arXiv:2502.01157}, 2025.

\bibitem{charatan2024pixelsplat}
David Charatan, Sizhe~Lester Li, Andrea Tagliasacchi, and Vincent Sitzmann.
\newblock pixelsplat: 3d gaussian splats from image pairs for scalable generalizable 3d reconstruction.
\newblock In {\em CVPR}, pages 19457--19467, 2024.

\bibitem{chen2024mvsplat}
Yuedong Chen, Haofei Xu, Chuanxia Zheng, Bohan Zhuang, Marc Pollefeys, Andreas Geiger, Tat-Jen Cham, and Jianfei Cai.
\newblock Mvsplat: Efficient 3d gaussian splatting from sparse multi-view images.
\newblock In {\em ECCV}, pages 370--386. Springer, 2024.

\bibitem{wewer2024latentsplat}
Christopher Wewer, Kevin Raj, Eddy Ilg, Bernt Schiele, and Jan~Eric Lenssen.
\newblock latentsplat: Autoencoding variational gaussians for fast generalizable 3d reconstruction.
\newblock In {\em ECCV}, pages 456--473. Springer, 2024.

\bibitem{wang2024freesplat}
Yunsong Wang, Tianxin Huang, Hanlin Chen, and Gim~Hee Lee.
\newblock Freesplat: Generalizable 3d gaussian splatting towards free view synthesis of indoor scenes.
\newblock {\em NeurIPS}, 37:107326--107349, 2024.

\bibitem{zhang2024gaussian}
Shengjun Zhang, Xin Fei, Fangfu Liu, Haixu Song, and Yueqi Duan.
\newblock Gaussian graph network: Learning efficient and generalizable gaussian representations from multi-view images.
\newblock {\em NeurIPS}, 37:50361--50380, 2024.

\bibitem{xu2024depthsplat}
Haofei Xu, Songyou Peng, Fangjinhua Wang, Hermann Blum, Daniel Barath, Andreas Geiger, and Marc Pollefeys.
\newblock Depthsplat: Connecting gaussian splatting and depth.
\newblock In {\em CVPR}, 2025.

\bibitem{ziwen2024long}
Chen Ziwen, Hao Tan, Kai Zhang, Sai Bi, Fujun Luan, Yicong Hong, Li~Fuxin, and Zexiang Xu.
\newblock Long-lrm: Long-sequence large reconstruction model for wide-coverage gaussian splats.
\newblock {\em arXiv preprint arXiv:2410.12781}, 2024.

\bibitem{reizenstein2021common}
Jeremy Reizenstein, Roman Shapovalov, Philipp Henzler, Luca Sbordone, Patrick Labatut, and David Novotny.
\newblock Common objects in 3d: Large-scale learning and evaluation of real-life 3d category reconstruction.
\newblock In {\em ICCV}, pages 10901--10911, 2021.

\bibitem{yeshwanth2023scannet++}
Chandan Yeshwanth, Yueh-Cheng Liu, Matthias Nie{\ss}ner, and Angela Dai.
\newblock Scannet++: A high-fidelity dataset of 3d indoor scenes.
\newblock In {\em ICCV}, pages 12--22, 2023.

\bibitem{yu2023mvimgnet}
Xianggang Yu, Mutian Xu, Yidan Zhang, Haolin Liu, Chongjie Ye, Yushuang Wu, Zizheng Yan, Chenming Zhu, Zhangyang Xiong, Tianyou Liang, et~al.
\newblock Mvimgnet: A large-scale dataset of multi-view images.
\newblock In {\em CVPR}, pages 9150--9161, 2023.

\bibitem{ling2024dl3dv}
Lu~Ling, Yichen Sheng, Zhi Tu, Wentian Zhao, Cheng Xin, Kun Wan, Lantao Yu, Qianyu Guo, Zixun Yu, Yawen Lu, et~al.
\newblock Dl3dv-10k: A large-scale scene dataset for deep learning-based 3d vision.
\newblock In {\em CVPR}, pages 22160--22169, 2024.

\bibitem{tishby2000information}
Naftali Tishby, Fernando~C Pereira, and William Bialek.
\newblock The information bottleneck method.
\newblock {\em arXiv preprint physics/0004057}, 1999.

\bibitem{dao2023flashattention}
Tri Dao.
\newblock Flashattention-2: Faster attention with better parallelism and work partitioning.
\newblock In {\em ICLR}, 2023.

\bibitem{chen2016training}
Tianqi Chen, Bing Xu, Chiyuan Zhang, and Carlos Guestrin.
\newblock Training deep nets with sublinear memory cost.
\newblock {\em CoRR}, 2016.

\bibitem{zhou2018stereo}
Tinghui Zhou, Richard Tucker, John Flynn, Graham Fyffe, and Noah Snavely.
\newblock Stereo magnification: learning view synthesis using multiplane images.
\newblock {\em ACM Transactions on Graphics}, 37(4):1--12, 2018.

\bibitem{dosovitskiy2020vit}
Alexey Dosovitskiy, Lucas Beyer, Alexander Kolesnikov, Dirk Weissenborn, Xiaohua Zhai, Thomas Unterthiner, Mostafa Dehghani, Matthias Minderer, Georg Heigold, Sylvain Gelly, Jakob Uszkoreit, and Neil Houlsby.
\newblock An image is worth 16x16 words: Transformers for image recognition at scale.
\newblock In {\em ICLR}, 2021.

\bibitem{darcet2023vitneedreg}
Timothée Darcet, Maxime Oquab, Julien Mairal, and Piotr Bojanowski.
\newblock Vision transformers need registers.
\newblock In {\em ICLR}, 2024.

\bibitem{caron2021emerging}
Mathilde Caron, Hugo Touvron, Ishan Misra, Herv\'e J\'egou, Julien Mairal, Piotr Bojanowski, and Armand Joulin.
\newblock Emerging properties in self-supervised vision transformers.
\newblock In {\em ICCV}, 2021.

\bibitem{oquab2023dinov2}
Maxime Oquab, Timothée Darcet, Theo Moutakanni, Huy~V. Vo, Marc Szafraniec, Vasil Khalidov, Pierre Fernandez, Daniel Haziza, Francisco Massa, Alaaeldin El-Nouby, Russell Howes, Po-Yao Huang, Hu~Xu, Vasu Sharma, Shang-Wen Li, Wojciech Galuba, Mike Rabbat, Mido Assran, Nicolas Ballas, Gabriel Synnaeve, Ishan Misra, Herve Jegou, Julien Mairal, Patrick Labatut, Armand Joulin, and Piotr Bojanowski.
\newblock Dinov2: Learning robust visual features without supervision.
\newblock In {\em ICLR}, 2025.

\bibitem{mahabadi2021variational}
Yonatan Belinkov, James Henderson, et~al.
\newblock Variational information bottleneck for effective low-resource fine-tuning.
\newblock In {\em ICLR}, 2020.

\bibitem{pmlr_v80_dai18d}
Bin Dai, Chen Zhu, Baining Guo, and David Wipf.
\newblock Compressing neural networks using the variational information bottleneck.
\newblock In Jennifer Dy and Andreas Krause, editors, {\em ICML}, volume~80, pages 1135--1144. PMLR, 2018.

\bibitem{Kim_2020_CVPR_Workshops}
Jinkyu Kim and Mayank Bansal.
\newblock Attentional bottleneck: Towards an interpretable deep driving network.
\newblock In {\em CVPR Workshops}, June 2020.

\bibitem{bai2025rethinking}
Shuanghao Bai, Wanqi Zhou, Pengxiang Ding, Wei Zhao, Donglin Wang, and Badong Chen.
\newblock Rethinking latent redundancy in behavior cloning: An information bottleneck approach for robot manipulation.
\newblock In {\em ICML}, 2025.

\bibitem{alemi2016deep}
Alexander~A Alemi, Ian Fischer, Joshua~V Dillon, and Kevin Murphy.
\newblock Deep variational information bottleneck.
\newblock In {\em ICLR}, 2017.

\bibitem{federici2020learningrobustrepresentationsmultiview}
Marco Federici, Anjan Dutta, Patrick Forr{\'e}, Nate Kushman, and Zeynep Akata.
\newblock Learning robust representations via multi-view information bottleneck.
\newblock In {\em ICLR}, 2020.

\bibitem{zhang2024discoveringcommoninformationmultiview}
Qi~Zhang, Mingfei Lu, Shujian Yu, Jingmin Xin, and Badong Chen.
\newblock Discovering common information in multi-view data.
\newblock {\em Information Fusion}, 108:102400, 2024.

\bibitem{xie2025gaussmigaussiansplattingshannon}
Yuhan Xie, Yixi Cai, Yinqiang Zhang, Lei Yang, and Jia Pan.
\newblock Gauss-mi: Gaussian splatting shannon mutual information for active 3d reconstruction.
\newblock {\em arXiv preprint arXiv:2504.21067}, 2025.

\bibitem{li2025streamgsonlinegeneralizablegaussian}
Yang Li, Jinglu Wang, Lei Chu, Xiao Li, Shiu-hong Kao, Ying-Cong Chen, and Yan Lu.
\newblock Streamgs: Online generalizable gaussian splatting reconstruction for unposed image streams.
\newblock {\em arXiv preprint arXiv:2503.06235}, 2025.

\bibitem{liu2024compgsefficient3dscene}
Xiangrui Liu, Xinju Wu, Pingping Zhang, Shiqi Wang, Zhu Li, and Sam Kwong.
\newblock Compgs: Efficient 3d scene representation via compressed gaussian splatting.
\newblock In {\em ACMMM}, pages 2936--2944, 2024.

\bibitem{fan2024lightgaussianunbounded3dgaussian}
Zhiwen Fan, Kevin Wang, Kairun Wen, Zehao Zhu, Dejia Xu, Zhangyang Wang, et~al.
\newblock Lightgaussian: Unbounded 3d gaussian compression with 15x reduction and 200+ fps.
\newblock {\em NeurIPS}, 37:140138--140158, 2024.

\bibitem{chen2023view}
Shenchang~Eric Chen and Lance Williams.
\newblock View interpolation for image synthesis.
\newblock In {\em Proceedings of the 20th Annual Conference on Computer Graphics and Interactive Techniques}, SIGGRAPH '93, page 279–288, New York, NY, USA, 1993. Association for Computing Machinery.

\bibitem{seitz1996view}
Steven~M Seitz and Charles~R Dyer.
\newblock View morphing.
\newblock In {\em Proceedings of the 23rd annual conference on Computer graphics and interactive techniques}, pages 21--30, 1996.

\bibitem{lombardi2019neural}
Stephen Lombardi, Tomas Simon, Jason Saragih, Gabriel Schwartz, Andreas Lehrmann, and Yaser Sheikh.
\newblock Neural volumes: learning dynamic renderable volumes from images.
\newblock {\em ACM Transactions on Graphics}, 38(4):1--14, 2019.

\bibitem{sitzmann2019scene}
Vincent Sitzmann, Michael Zollh{\"o}fer, and Gordon Wetzstein.
\newblock Scene representation networks: Continuous 3d-structure-aware neural scene representations.
\newblock {\em NeurIPS}, 32, 2019.

\bibitem{sitzmann2021light}
Vincent Sitzmann, Semon Rezchikov, Bill Freeman, Josh Tenenbaum, and Fredo Durand.
\newblock Light field networks: Neural scene representations with single-evaluation rendering.
\newblock {\em NeurIPS}, 34:19313--19325, 2021.

\bibitem{mildenhall2020nerf}
Ben Mildenhall, Pratul~P Srinivasan, Matthew Tancik, Jonathan~T Barron, Ravi Ramamoorthi, and Ren Ng.
\newblock Nerf: Representing scenes as neural radiance fields for view synthesis.
\newblock {\em Communications of the ACM}, 65(1):99--106, 2021.

\bibitem{wang2022r2l}
Huan Wang, Jian Ren, Zeng Huang, Kyle Olszewski, Menglei Chai, Yun Fu, and Sergey Tulyakov.
\newblock R2l: Distilling neural radiance field to neural light field for efficient novel view synthesis.
\newblock In {\em ECCV}, 2022.

\bibitem{sun2021direct}
Cheng Sun, Min Sun, and Hwann-Tzong Chen.
\newblock Direct voxel grid optimization: Super-fast convergence for radiance fields reconstruction.
\newblock In {\em CVPR}, 2022.

\bibitem{kangle2021dsnerf}
Kangle Deng, Andrew Liu, Jun-Yan Zhu, and Deva Ramanan.
\newblock Depth-supervised nerf: Fewer views and faster training for free.
\newblock In {\em CVPR}, pages 12882--12891, 2022.

\bibitem{muller2022instant}
Thomas M{\"u}ller, Alex Evans, Christoph Schied, and Alexander Keller.
\newblock Instant neural graphics primitives with a multiresolution hash encoding.
\newblock {\em ACM transactions on graphics}, 41(4):1--15, 2022.

\bibitem{chen2022tensorf}
Anpei Chen, Zexiang Xu, Andreas Geiger, Jingyi Yu, and Hao Su.
\newblock Tensorf: Tensorial radiance fields.
\newblock In {\em ECCV}, pages 333--350. Springer, 2022.

\bibitem{cao2024lightning}
Junyi Cao, Zhichao Li, Naiyan Wang, and Chao Ma.
\newblock Lightning nerf: Efficient hybrid scene representation for autonomous driving.
\newblock In {\em ICRA}, pages 16803--16809. IEEE, 2024.

\bibitem{kajiya1984ray}
James~T Kajiya and Brian~P Von~Herzen.
\newblock Ray tracing volume densities.
\newblock {\em ACM SIGGRAPH computer graphics}, 18(3):165--174, 1984.

\bibitem{fan2024instantsplat}
Zhiwen Fan, Wenyan Cong, Kairun Wen, Kevin Wang, Jian Zhang, Xinghao Ding, Danfei Xu, Boris Ivanovic, Marco Pavone, Georgios Pavlakos, et~al.
\newblock Instantsplat: Unbounded sparse-view pose-free gaussian splatting in 40 seconds.
\newblock {\em CoRR}, 2024.

\bibitem{lu2024scaffold}
Tao Lu, Mulin Yu, Linning Xu, Yuanbo Xiangli, Limin Wang, Dahua Lin, and Bo~Dai.
\newblock Scaffold-gs: Structured 3d gaussians for view-adaptive rendering.
\newblock In {\em CVPR}, pages 20654--20664, 2024.

\bibitem{girish2024eagles}
Sharath Girish, Kamal Gupta, and Abhinav Shrivastava.
\newblock Eagles: Efficient accelerated 3d gaussians with lightweight encodings.
\newblock In {\em ECCV}, pages 54--71. Springer, 2024.

\bibitem{yu2021pixelnerfneuralradiancefields}
Alex Yu, Vickie Ye, Matthew Tancik, and Angjoo Kanazawa.
\newblock pixelnerf: Neural radiance fields from one or few images.
\newblock In {\em CVPR}, pages 4578--4587, 2021.

\bibitem{wang2021ibrnetlearningmultiviewimagebased}
Qianqian Wang, Zhicheng Wang, Kyle Genova, Pratul~P Srinivasan, Howard Zhou, Jonathan~T Barron, Ricardo Martin-Brualla, Noah Snavely, and Thomas Funkhouser.
\newblock Ibrnet: Learning multi-view image-based rendering.
\newblock In {\em CVPR}, pages 4690--4699, 2021.

\bibitem{chen2021mvsnerffastgeneralizableradiance}
Anpei Chen, Zexiang Xu, Fuqiang Zhao, Xiaoshuai Zhang, Fanbo Xiang, Jingyi Yu, and Hao Su.
\newblock Mvsnerf: Fast generalizable radiance field reconstruction from multi-view stereo.
\newblock In {\em ICCV}, pages 14124--14133, 2021.

\bibitem{johari2022geonerfgeneralizingnerfgeometry}
Mohammad~Mahdi Johari, Yann Lepoittevin, and Fran{\c{c}}ois Fleuret.
\newblock Geonerf: Generalizing nerf with geometry priors.
\newblock In {\em CVPR}, pages 18365--18375, 2022.

\bibitem{chen2023explicit}
Yuedong Chen, Haofei Xu, Qianyi Wu, Chuanxia Zheng, Tat-Jen Cham, and Jianfei Cai.
\newblock Explicit correspondence matching for generalizable neural radiance fields.
\newblock {\em IEEE Transactions on Pattern Analysis and Machine Intelligence}, 2025.

\bibitem{xu2024murfmultibaselineradiancefields}
Haofei Xu, Anpei Chen, Yuedong Chen, Christos Sakaridis, Yulun Zhang, Marc Pollefeys, Andreas Geiger, and Fisher Yu.
\newblock Murf: multi-baseline radiance fields.
\newblock In {\em CVPR}, pages 20041--20050, 2024.

\bibitem{fei2024pixelgaussian}
Xin Fei, Wenzhao Zheng, Yueqi Duan, Wei Zhan, Masayoshi Tomizuka, Kurt Keutzer, and Jiwen Lu.
\newblock Pixelgaussian: Generalizable 3d gaussian reconstruction from arbitrary views.
\newblock {\em arXiv preprint arXiv:2410.18979}, 2024.

\bibitem{ye2024no}
Botao Ye, Sifei Liu, Haofei Xu, Xueting Li, Marc Pollefeys, Ming-Hsuan Yang, and Songyou Peng.
\newblock No pose, no problem: Surprisingly simple 3d gaussian splats from sparse unposed images.
\newblock {\em arXiv preprint arXiv:2410.24207}, 2024.

\bibitem{min2024epipolarfree}
Zhiyuan Min, Yawei Luo, Jianwen Sun, and Yi~Yang.
\newblock Epipolar-free 3d gaussian splatting for generalizable novel view synthesis.
\newblock In A.~Globerson, L.~Mackey, D.~Belgrave, A.~Fan, U.~Paquet, J.~Tomczak, and C.~Zhang, editors, {\em NeurIPS}, volume~37, pages 39573--39596. Curran Associates, Inc., 2024.

\bibitem{kang2025selfsplat}
Gyeongjin Kang, Jisang Yoo, Jihyeon Park, Seungtae Nam, Hyeonsoo Im, Sangheon Shin, Sangpil Kim, and Eunbyung Park.
\newblock Selfsplat: Pose-free and 3d prior-free generalizable 3d gaussian splatting.
\newblock In {\em CVPR}, pages 22012--22022, 2025.

\bibitem{chen2024mvsplat360}
Yuedong Chen, Chuanxia Zheng, Haofei Xu, Bohan Zhuang, Andrea Vedaldi, Tat-Jen Cham, and Jianfei Cai.
\newblock Mvsplat360: Feed-forward 360 scene synthesis from sparse views.
\newblock {\em NeurIPS}, 37:107064--107086, 2024.

\bibitem{schoenberger2016sfm}
Johannes~Lutz Sch\"{o}nberger and Jan-Michael Frahm.
\newblock Structure-from-motion revisited.
\newblock In {\em CVPR}, 2016.

\bibitem{schoenberger2016mvs}
Johannes~Lutz Sch\"{o}nberger, Enliang Zheng, Marc Pollefeys, and Jan-Michael Frahm.
\newblock Pixelwise view selection for unstructured multi-view stereo.
\newblock In {\em ECCV}, 2016.

\bibitem{wang2004image}
Zhou Wang, Alan~C Bovik, Hamid~R Sheikh, and Eero~P Simoncelli.
\newblock Image quality assessment: from error visibility to structural similarity.
\newblock {\em IEEE transactions on image processing}, 13(4):600--612, 2004.

\bibitem{zhang2018unreasonable}
Richard Zhang, Phillip Isola, Alexei~A Efros, Eli Shechtman, and Oliver Wang.
\newblock The unreasonable effectiveness of deep features as a perceptual metric.
\newblock In {\em CVPR}, pages 586--595, 2018.

\bibitem{liu2021infinite}
Andrew Liu, Richard Tucker, Varun Jampani, Ameesh Makadia, Noah Snavely, and Angjoo Kanazawa.
\newblock Infinite nature: Perpetual view generation of natural scenes from a single image.
\newblock In {\em ICCV}, pages 14458--14467, 2021.

\bibitem{xiong2020layer}
Ruibin Xiong, Yunchang Yang, Di~He, Kai Zheng, Shuxin Zheng, Chen Xing, Huishuai Zhang, Yanyan Lan, Liwei Wang, and Tieyan Liu.
\newblock On layer normalization in the transformer architecture.
\newblock In {\em International conference on machine learning}, pages 10524--10533. PMLR, 2020.

\bibitem{dao2022flashattention}
Tri Dao, Dan Fu, Stefano Ermon, Atri Rudra, and Christopher R{\'e}.
\newblock Flashattention: Fast and memory-efficient exact attention with io-awareness.
\newblock {\em NeurIPS}, 35:16344--16359, 2022.

\bibitem{loshchilov2017decoupled}
Ilya Loshchilov and Frank Hutter.
\newblock Decoupled weight decay regularization.
\newblock In {\em ICLR}, 2019.

\end{thebibliography}
\normalsize

\newpage
\appendix
\setcounter{table}{0}
\setcounter{figure}{0}
\setcounter{section}{0}
\renewcommand{\thetable}{\Alph{table}}
\renewcommand{\thefigure}{\Alph{figure}}
\renewcommand{\thesection}{\Alph{section}} 
\section{More Experimental Analysis}
\label{sec:app_experiment}

\subsection{Cross Dataset Generalization} 
Following MVSplat~\cite{chen2024mvsplat}, we conducted experiments using a pretrained model on the RealEstate10K (RE10K) dataset~\cite{zhou2018stereo} (as detailed in \cref{tab:multi view results re10k}) and tested its performance on the ACID dataset~\cite{liu2021infinite} to evaluate the generalization capabilities of our proposed \method across diverse datasets. As demonstrated in Table \ref{tab:cross_acid_gen}, MVSplat with \method exhibits remarkable efficacy in cross-dataset generalization. Notably, this performance advantage becomes progressively more pronounced with an increasing number of input views.

\begin{table}[htbp]
    \centering
    \caption{\textbf{Quantitative comparison on ACID~\cite{liu2021infinite} with trained model on RE10K.} Trained on indoor scenes (RE10K), MVSplat~\cite{chen2024mvsplat} and pixelSplat~\cite{charatan2024pixelsplat} with \method perform much better as evaluated on the ACID dataset.} %
    \vspace{0.2cm}
    \setlength{\tabcolsep}{2mm}{\begin{tabular}{llccc} %
        \toprule
        Views & Methods & PSNR$\uparrow$ & SSIM$\uparrow$ & LPIPS$\downarrow$ \\
        \midrule
        \multirow{4}{*}{36 views}& pixelSplat & OOM & OOM & OOM \\
        & $\text{pixelSplat + Ours}$ & \textbf{27.78} & \textbf{0.823} & \textbf{0.238} \\
        & $\text{MVSplat}$ & 24.89 & 0.812 & 0.179 \\
        & $\text{MVSplat + Ours}$ & \phantom{\bf \tiny +0.01}\hfill \textbf{28.16}{\raggedleft \color{soft_red} \bf \tiny +3.27} & \phantom{\bf \tiny +0.001}\hfill \textbf{0.853}{\raggedleft \color{soft_red} \bf \tiny +0.041} & \phantom{\bf \tiny +0.001}\hfill \textbf{0.145}{\raggedleft \color{soft_red} \bf \tiny -0.034} \\
        \midrule
        \multirow{4}{*}{24 views}& pixelSplat & OOM & OOM & OOM \\
        & $\text{pixelSplat + Ours}$ & \textbf{27.91} & \textbf{0.825} & \textbf{0.235}\\
        & $\text{MVSplat}$ & 25.46 & 0.829 & 0.167 \\
        & $\text{MVSplat + Ours}$ & \phantom{\bf \tiny +0.01}\hfill \textbf{28.33}{\raggedleft \color{soft_red} \bf \tiny +2.87} & \phantom{\bf \tiny +0.001}\hfill \textbf{0.856}{\raggedleft \color{soft_red} \bf \tiny +0.027} & \phantom{\bf \tiny +0.001}\hfill \textbf{0.142}{\raggedleft \color{soft_red} \bf \tiny -0.025} \\
        \midrule
        \multirow{4}{*}{16 views}& pixelSplat & OOM & OOM & OOM \\
        & $\text{pixelSplat + Ours}$ & \textbf{27.97} & \textbf{0.826} & \textbf{0.234} \\
        & $\text{MVSplat}$ & 26.08 & 0.844 & 0.156 \\
        & $\text{MVSplat + Ours}$ & \phantom{\bf \tiny +0.01}\hfill \textbf{28.42}{\raggedleft \color{soft_red} \bf \tiny +2.34} & \phantom{\bf \tiny +0.001}\hfill \textbf{0.858}{\raggedleft \color{soft_red} \bf \tiny +0.014} & \phantom{\bf \tiny +0.001}\hfill \textbf{0.141}{\raggedleft \color{soft_red} \bf \tiny -0.015} \\
        \midrule
        \multirow{4}{*}{8 views}& pixelSplat & 26.69 & 0.807 & 0.260\\
        & $\text{pixelSplat + Ours}$ & \phantom{\bf \tiny +0.01}\hfill \textbf{28.05}{\raggedleft \color{soft_red} \bf \tiny +1.36} & \phantom{\bf \tiny +0.001}\hfill \textbf{0.828}{\raggedleft \color{soft_red} \bf \tiny +0.021} & \phantom{\bf \tiny +0.001}\hfill \textbf{0.234}{\raggedleft \color{soft_red} \bf \tiny -0.026}\\
        & $\text{MVSplat}$ & 27.89 & \textbf{0.864} & \textbf{0.140} \\
        & $\text{MVSplat + Ours}$ & \phantom{\bf \tiny +0.01}\hfill \textbf{28.60}{\raggedleft \color{soft_red} \bf \tiny +0.71} & \phantom{\bf \tiny +0.001}\hfill 0.860{\raggedleft \color{dark_green} \bf \tiny -0.004} & \phantom{\bf \tiny +0.001}\hfill \textbf{0.140}{\raggedleft \color{soft_red} \bf \tiny -0.000} \\ %
        \bottomrule
    \end{tabular}}
    \vspace{-0.3cm}
    \label{tab:cross_acid_gen}
\end{table}

\subsection{Other Selection Strategies} 

In addition to our final choice of FPS-based selection strategy (\cref{fig:pipeline} Step~1), we also experimented with other anchor views selection strategies. Below are detailed implementation descriptions for each strategy and a comparison of all strategies' performance in \cref{tab:selection_strategies}.

\boldstart{Overlap-based selection.}
We try overlap-based selection by computing a pairwise overlap matrix between camera views. The overlap is obtained by projecting dense pixel rays from one view into another and averaging the proportion of pixels that fall within the target image plane. The final overlap score between two views is the minimum of the bidirectional projections. We then apply a greedy vectorized algorithm that iteratively selects views with the lowest average overlap to the already chosen set, ensuring maximal scene coverage with minimal redundancy. This procedure produces both the anchor indices and the cluster assignments for the remaining views.


\boldstart{Pose-free settings.}
In the absence of camera poses, we cluster learned per-view tokens instead of 3D positions. We use K-Means in the feature space of global view embeddings. Each token is assigned to its nearest cluster center, and the closest token to the mean of each cluster is selected as the anchor. This pose-free clustering provides a practical alternative for unposed datasets, as the anchor set is derived entirely from image content rather than camera geometry.

\begin{table}[htbp]
    \centering
    \caption{\textbf{Quantitative comparison of different selection strategies integrated with DepthSplat~\cite{xu2024depthsplat}.} All variants are evaluated with 36 test views on DL3DV~\cite{ling2024dl3dv}.}
    \vspace{0.2cm}
    \setlength{\tabcolsep}{2.5mm}
    {\begin{tabular}{lccc}
        \toprule
        Methods & PSNR$\uparrow$ & SSIM$\uparrow$ & LPIPS$\downarrow$ \\
        \midrule
        DepthSplat 
            & \phantom{\bf \tiny +0.01}\hfill 19.23\phantom{\bf \tiny +0.01}
            & \phantom{\bf \tiny +0.01}\hfill 0.666\phantom{\bf \tiny +0.01} 
            & \phantom{\bf \tiny +0.01}\hfill 0.286\phantom{\bf \tiny +0.01} \\
        DepthSplat + Ours (overlap-based) 
            & \phantom{\bf \tiny +0.01}\hfill 21.49{\raggedleft \color{soft_red} \tiny +2.26} 
            & \phantom{\bf \tiny +0.001}\hfill 0.727{\raggedleft \color{soft_red} \tiny +0.061} 
            & \phantom{\bf \tiny +0.001}\hfill 0.194{\raggedleft \color{soft_red} \tiny -0.092} \\
        DepthSplat + Ours (pose-free) 
            & \phantom{\bf \tiny +0.01}\hfill 22.81{\raggedleft \color{soft_red} \tiny +3.58} 
            & \phantom{\bf \tiny +0.001}\hfill 0.791{\raggedleft \color{soft_red} \tiny +0.125} 
            & \phantom{\bf \tiny +0.001}\hfill 0.174{\raggedleft \color{soft_red} \tiny -0.112} \\
        DepthSplat + Ours (K-Means-based) 
            & \phantom{\bf \tiny +0.01}\hfill 22.84{\raggedleft \color{soft_red} \tiny +3.61} 
            & \phantom{\bf \tiny +0.001}\hfill 0.789{\raggedleft \color{soft_red} \tiny +0.123} 
            & \phantom{\bf \tiny +0.001}\hfill 0.175{\raggedleft \color{soft_red} \tiny -0.111} \\
        DepthSplat + Ours
            & \phantom{\bf \tiny +0.01}\hfill \textbf{23.88}{\raggedleft \color{soft_red} \bf \tiny +4.65} 
            & \phantom{\bf \tiny +0.001}\hfill \textbf{0.815}{\raggedleft \color{soft_red} \bf \tiny +0.149} 
            & \phantom{\bf \tiny +0.001}\hfill \textbf{0.150}{\raggedleft \color{soft_red} \bf \tiny -0.136} \\
        \bottomrule
    \end{tabular}}
    \vspace{-0.3cm}
    \label{tab:selection_strategies}
\end{table}

\boldstart{K-Means-based selection.}
K-Means-based selection groups view positions directly. Given the 3D camera centers of all input views, we apply K-Means with a fixed number of groups and identify the anchor of each cluster as the view closest to its centroid. Each remaining view is assigned to its nearest centroid, producing compact and spatially coherent groups. This procedure is fully deterministic given the random seed and provides stable anchors that reflect the geometric distribution of input cameras.

\subsection{Comparison with Confidence-based Pruning}

We compare \method against confidence-based pruning after predicting Gaussians in MVSplat~\cite{chen2024mvsplat}. 
Confidence pruning removes input views based on their predicted reliability scores, with a fixed pruning ratio controlling the proportion of views retained. 
When maintaining the same number of Gaussians, the aggressive removal of views leads to a substantial performance drop due to insufficient multi-view support. 
At a pruning ratio of $0.5$, the method achieve the best accuracy, but the gain remains limited because the retained subset does not guarantee spatial coverage of the scene. 
In contrast, \method consistently compresses all input views into compact latent anchors while maintaining balanced coverage, which yields superior accuracy without discarding information.

\begin{table}[htbp]
    \centering
    \caption{\textbf{Comparison with confidence-based pruning (CP) in MVSplat~\cite{chen2024mvsplat} under 24 views on RE10K.} For a fair comparison, the prune ratio for CP is set to $0.75$ to yield a same number of Gaussians as \method, and we also manually adjust the prune ratio to $0.5$ to achieve the best performance. All models are re-trained.}
    \vspace{0.2cm}
    \setlength{\tabcolsep}{2.5mm}
    {\begin{tabular}{lccc}
        \toprule
        Methods & PSNR$\uparrow$ & SSIM$\uparrow$ & LPIPS$\downarrow$ \\
        \midrule
        MVSplat 
            & \phantom{\bf \tiny +0.01}\hfill 25.00\phantom{\bf \tiny +0.01}
            & \phantom{\bf \tiny +0.001}\hfill 0.871\phantom{\bf \tiny +0.001} 
            & \phantom{\bf \tiny +0.001}\hfill 0.137\phantom{\bf \tiny +0.001} \\
        MVSplat + CP ($prune\_ratio=0.75$) 
            & \phantom{\bf \tiny +0.01}\hfill 21.15{\raggedleft \color{dark_green} \tiny -3.85} 
            & \phantom{\bf \tiny +0.001}\hfill 0.816{\raggedleft \color{dark_green} \tiny -0.055} 
            & \phantom{\bf \tiny +0.001}\hfill 0.190{\raggedleft \color{dark_green} \tiny +0.053} \\
        MVSplat + CP ($prune\_ratio=0.5$) 
            & \phantom{\bf \tiny +0.01}\hfill 26.94{\raggedleft \color{soft_red} \tiny +1.94} 
            & \phantom{\bf \tiny +0.001}\hfill 0.886{\raggedleft \color{soft_red} \tiny +0.015} 
            & \phantom{\bf \tiny +0.001}\hfill 0.130{\raggedleft \color{soft_red} \tiny -0.007} \\
        MVSplat + Ours 
            & \phantom{\bf \tiny +0.01}\hfill \textbf{27.49}{\raggedleft \color{soft_red} \bf \tiny +2.49} 
            & \phantom{\bf \tiny +0.001}\hfill \textbf{0.895}{\raggedleft \color{soft_red} \bf \tiny +0.024} 
            & \phantom{\bf \tiny +0.001}\hfill \textbf{0.111}{\raggedleft \color{soft_red} \bf \tiny -0.026} \\
        \bottomrule
    \end{tabular}}
    \vspace{-0.3cm}
    \label{tab:mvsplat_conf_pruning}
\end{table}

\subsection{Ablation Study of IB-Loss}

We conduct an ablation study to evaluate the role of the compression term derived from the Information Bottleneck formulation. 
When the compression term is removed ($\beta=0$), \method still improves significantly over the baseline DepthSplat, confirming that the architectural design alone provides strong benefits. 
Introducing a small but non-zero coefficient ($\beta=10^{-5}$) further encourages compact latent representations and yields the best balance between distortion and fidelity. 
This demonstrates that the IB-inspired loss can serve as a lightweight regularizer to assist compression module learning in achieving better compression results.

\begin{table}[htbp]
    \centering
    \caption{\textbf{Ablation of the IB-inspired compression loss on DepthSplat~\cite{xu2024depthsplat} with 36 views on DL3DV~\cite{ling2024dl3dv}.} The compression term provides additional regularization that improves overall performance.}
    \vspace{0.2cm}
    \setlength{\tabcolsep}{2.5mm}
    {\begin{tabular}{lccc}
        \toprule
        Methods & PSNR$\uparrow$ & SSIM$\uparrow$ & LPIPS$\downarrow$ \\
        \midrule
        DepthSplat 
            & \phantom{\bf \tiny +0.01}\hfill 19.23\phantom{\bf \tiny +0.01}
            & \phantom{\bf \tiny +0.001}\hfill 0.666\phantom{\bf \tiny +0.001} 
            & \phantom{\bf \tiny +0.001}\hfill 0.286\phantom{\bf \tiny +0.001} \\
        DepthSplat + Ours ($\beta=0$) 
            & \phantom{\bf \tiny +0.01}\hfill 23.43{\raggedleft \color{soft_red} \tiny +4.20} 
            & \phantom{\bf \tiny +0.001}\hfill 0.806{\raggedleft \color{soft_red} \tiny +0.140} 
            & \phantom{\bf \tiny +0.001}\hfill 0.165{\raggedleft \color{soft_red} \tiny -0.121} \\
        DepthSplat + Ours ($\beta=10^{-5}$) 
            & \phantom{\bf \tiny +0.01}\hfill \textbf{23.88}{\raggedleft \color{soft_red} \bf \tiny +4.65} 
            & \phantom{\bf \tiny +0.001}\hfill \textbf{0.815}{\raggedleft \color{soft_red} \bf \tiny +0.149} 
            & \phantom{\bf \tiny +0.001}\hfill \textbf{0.150}{\raggedleft \color{soft_red} \bf \tiny -0.136} \\
        \bottomrule
    \end{tabular}}
    \vspace{-0.3cm}
    \label{tab:ib_loss_ablation}
\end{table}

\section{More Implementation Details}
\label{sec:app_implementation}

\boldstart{Network architectures.} 
In \cref{alg:overall}, we provide a detailed description of how \method is integrated into existing feed-forward 3D Gaussian Splatting (3DGS) frameworks~\cite{charatan2024pixelsplat, xu2024depthsplat, chen2024mvsplat}. Initially, we select anchor views and their corresponding support views following \cref{alg:fps} and Eq. (5). The features associated with these views are then processed by an attention-based network. This network is composed of 6 structurally identical blocks, wherein each block encompasses a cross-attention layer, a self-attention layer, and an MLP layer. The cross-attention mechanism operates by employing the anchor features as query, while the support features provide the key and value. Subsequent to this fusion, the resulting features are further refined by the self-attention and MLP layers.

\begin{algorithm}
    \caption{Overview of Feed-Forward 3DGS framework with \method.}
    \label{alg:overall}
    \renewcommand{\algorithmicrequire}{\textbf{Input:}}
    \renewcommand{\algorithmicensure}{\textbf{Output:}}

    \begin{algorithmic}
        \Require $K$ input views $\mathcal{V}=\{V_{i}\}_{i=1}^{K}$, camera poses $\mathcal{P}=\{P_{i}\}_{i=1}^{K}$, the number of anchor views $N$, the number of network blocks $h$.
        \Ensure Gaussian parameters $\mathcal{Y}=\{(\mu,\Sigma,\alpha,c)\}$.
        \State $\mathcal{X}\leftarrow\Phi_{image}(\mathcal{V}, \mathcal{P})$
        \State $\mathcal{X}_\mathrm{anchor}, \mathcal{X}_\mathrm{support}\leftarrow \mathcal{X}$, with Anchor view selection.
        \State Assign support views to anchor cluster $\mathcal{C}\leftarrow \mathcal{X}_\mathrm{support}$
        \State Initialize state $\mathcal{Z}\leftarrow\mathcal{X}_\mathrm{anchor}$
        \For{$i \gets 1$ to $h$}
            \State $\mathcal{Z}\leftarrow \text{Attention}(Q, K, V)$, where $ Q \leftarrow \mathcal{Z} \quad K, V \leftarrow \mathcal{X}_\mathrm{support}$
            \State $\mathcal{Z}\leftarrow\text{Attention}(Q, K, V)$, where $Q, K, V \leftarrow \mathcal{Z}$
            \State $\mathcal{Z}\leftarrow\text{MLP}(\mathcal{Z})$
        \EndFor
        \State $\{(\mathbf{\mu}_{i}, \mathbf{\Sigma}_{i}, \mathbf{\alpha}_{i}, \mathbf{c}_{i})\} \leftarrow\Psi_\mathrm{pred}(\mathcal{Z}, \mathcal{P})$
        \State \Return $\mathcal{Y} \leftarrow \{(\mathbf{\mu}_{i}, \mathbf{\Sigma}_{i}, \mathbf{\alpha}_{i}, \mathbf{c}_{i})\}$
    \end{algorithmic}
\end{algorithm}

\begin{algorithm}
    \caption{Farthest Point Sampling for Anchor View Selection}
    \label{alg:fps}
    \renewcommand{\algorithmicrequire}{\textbf{Input:}}
    \renewcommand{\algorithmicensure}{\textbf{Output:}}

    \begin{algorithmic}
        \Require Set of view camera positions $\mathcal{T} = \{\mathbf{T}_1, \mathbf{T}_2, ..., \mathbf{T}_K\}$, Number of anchor views $N$
        \Ensure Indices of the selected anchor views $\mathcal{S} = \{\mathbf{T}_{a_1}, \mathbf{T}_{a_2}, ..., \mathbf{T}_{a_n}\}$    
        \State Initialize the set of anchor view indices $\mathcal{S} \gets \emptyset$
        \State Randomly select a random anchor view $\mathbf{T}_{a_1} \in \mathcal{T}$, where $\mathbf{T}_{a_1} \sim \text{Uniform}(\mathcal{T})$
        \State Add $\mathbf{T}_{a_1}$ to $\mathcal{S}$: $\mathcal{S} \gets \{\mathbf{T}_{a_1}\}$
        \For{$j \gets 2$ to $N$}
            \State Initialize a dictionary to store minimum distances $D \gets \{\}$
            \For{$k \gets 1$ to $K$}
                \If{$k \notin \mathcal{S}$}
                    \State Calculate the minimum distance $d_k \gets \min_{i \in \mathcal{S}} \|\mathbf{T}_k - \mathbf{T}_i\|_2$
                    \State Store the distance: $D[k] \gets d_k$
                \EndIf
            \EndFor
            \State Find the view position $T_{a_j}$ with the maximum minimum distance: $T_{a_j} \gets \arg\max_{k \notin \mathcal{S}} D[k]$
            \State Add $a_j$ to $\mathcal{S}$: $\mathcal{S} \gets \mathcal{S} \cup \{T_{a_j}\}$
        \EndFor
        \State \Return $\mathcal{S}$
    \end{algorithmic}
\end{algorithm}

\begin{figure}
    \centering
    \includegraphics[width=\linewidth]{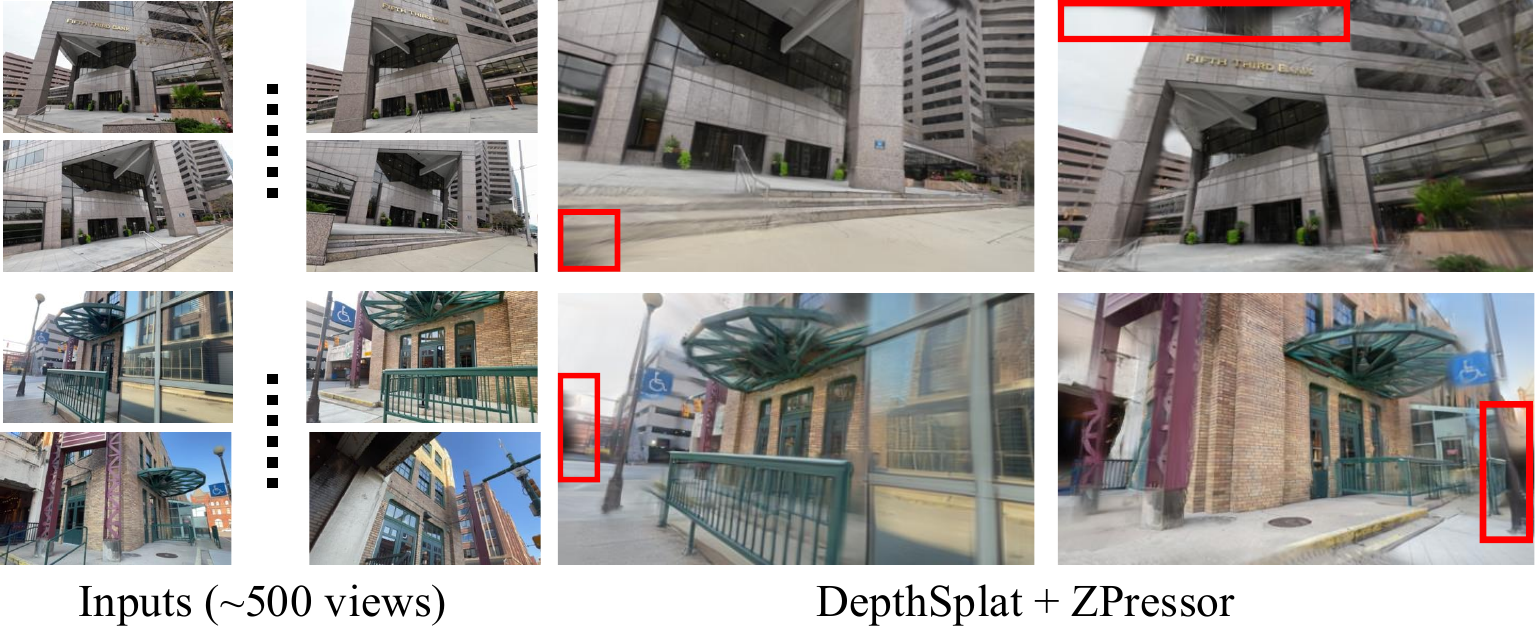}
    \caption{\textbf{Limitations.} Visual results from extremely dense input views show slightly poor presentation effect.}
    \label{fig:limit}
\end{figure}

To ensure training stability, deviating from traditional Transformer architectures, we employ Pre-Layer Normalization~\cite{xiong2020layer} (Pre-LN), which enhances the robustness of the model. Furthermore, system-level advancements have been incorporated to accelerate computation. For example, we employ FlashAttention~\cite{dao2022flashattention, dao2023flashattention}, which uses highly optimized GPU kernels and leverages hardware topology to compute attention in a time- and memory-efficient manner.

\boldstart{More training details.}
We use the first model version of DepthSplat~\cite{xu2024depthsplat} from its October 2024 release. Experimental results obtained with this specific version may exhibit slight variations when compared to the current version.
\method was incorporated subsequent to the monocular feature extraction performed by CNN. 

Adhering to its original configuration, experiments were conducted at a resolution of $256 \times 448$. The model was initially trained on the RE10K~\cite{zhou2018stereo} for 100,000 steps and subsequently fine-tuned on the DL3DV~\cite{ling2024dl3dv} for an additional 100,000 steps. We employed the AdamW optimizer~\cite{loshchilov2017decoupled} with a learning rate of $2 \times 10^{-4}$. The total training duration was approximately two days, and the integration of \method did not significantly alter the original training time of DepthSplat.

Similarly, for MVSplat~\cite{chen2024mvsplat} and pixelSplat~\cite{charatan2024pixelsplat}, \method was integrated after the monocular feature extraction stage. MVSplat utilizes a CNN for feature extraction, whereas pixelSplat employs DINO~\cite{caron2021emerging, oquab2023dinov2}; this architectural choice in pixelSplat contributes to a marginally higher VRAM consumption compared to the other two baselines. We maintained the model parameter settings as published in their respective original works, training models on the RE10K~\cite{zhou2018stereo} at a resolution of $256 \times 256$. The learning rate was set to $2 \times 10^{-4}$ for MVSplat and $1.5 \times 10^{-4}$ for pixelSplat, where both of which were trained for 100,000 steps. Notably, due to memory constraints, we trained the pixelSplat model incorporating \method using 4 anchor views, in contrast to the 6 anchor views configured for DepthSplat and MVSplat. The training times for MVSplat and pixelSplat, when augmented with \method, remained comparable to their original durations.

We will \textbf{open-source} the complete codebase for \method, our \method-integrated versions of DepthSplat, MVSplat, and pixelSplat, and all associated model checkpoints.

\section{Limitation and Societal Impacts}
\label{sec:limit_scoietal}

\boldstart{Limitation analysis.} 
As discussed in \cref{sec:conclusion}, \method exhibits limitations when processing scenarios with an extremely high density of input views. Specifically, its efficacy in compressing the information from such dense views through a limited set of anchor views is diminished. To illustrate this, we conducted an experiment on DepthSplat~\cite{xu2024depthsplat} integrated with \method, using approximately 500 images as input. As depicted in \cref{fig:limit}, the quality of the rendered novel views was perceptibly affected, which can be attributed to an insufficient number of Gaussian primitives to adequately represent the scene under these dense input conditions.

\boldstart{Potential and negative societal impacts.} 
\method can significantly reduce the training costs associated with feed-forward 3DGS networks. It enables the processing of a larger number of input views within the same VRAM budget and training duration, delivering high-fidelity rendering results and thereby decreasing energy consumption during the model training process.
While the capability to render higher-quality novel views from more densely sampled perspectives positions \method as a valuable tool for augmented reality applications, it is important to acknowledge that the fidelity of the rendering can be compromised by the emergence of artifacts, particularly when processing input views of extremely high density. Consequently, in safety-critical applications, such as the training of autonomous driving models, the deployment of \method would necessitate the implementation of additional precautionary measures to mitigate potential risks arising from such limitations.

\section{More Visual Comparisons}
\label{sec:app_visual}

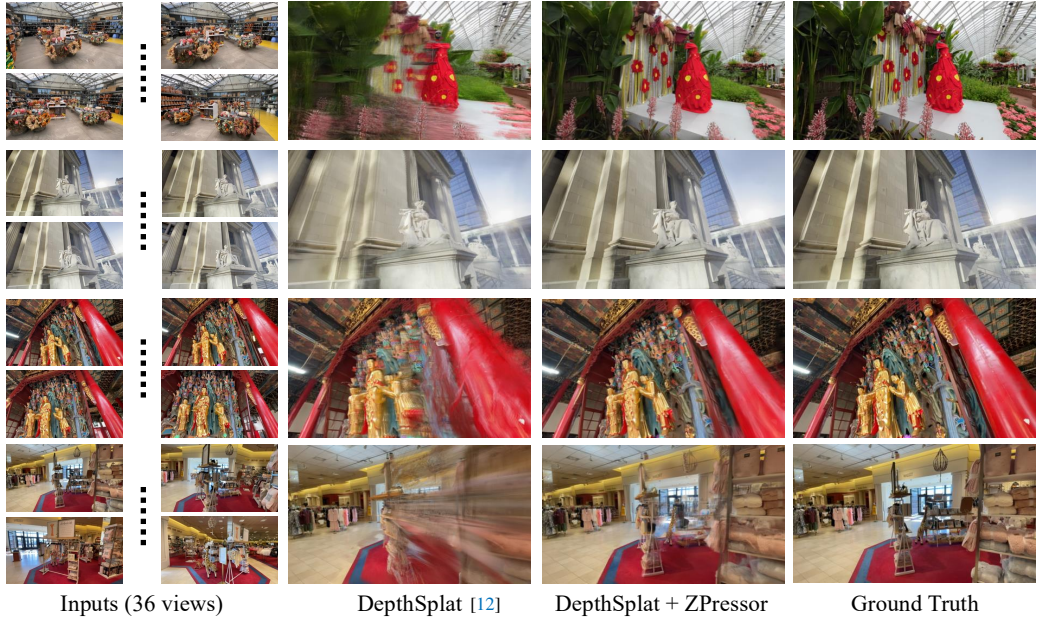
\begin{figure}
    \centering
    \includegraphics[width=\linewidth]{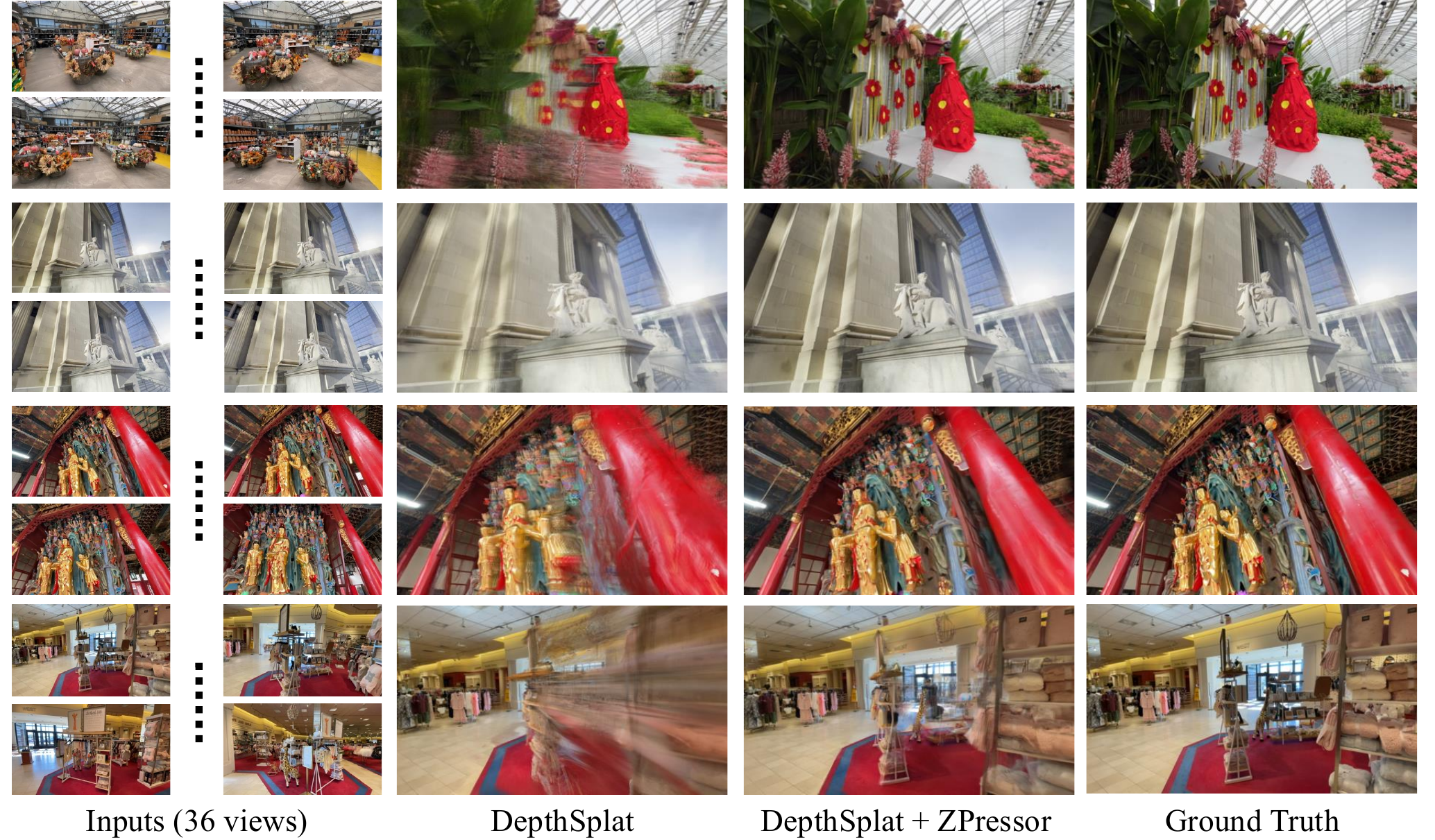}
    \begin{picture}(0,0)
    \put(-20,13.5){\scriptsize \cite{xu2024depthsplat}}
    \end{picture}
    \caption{\textbf{More qualitative comparisons on DL3DV~\cite{ling2024dl3dv} with DepthSplat~\cite{xu2024depthsplat} under 36 input views.} Models with \method performs the best in all cases.}
    \label{fig:more_dl3dv_36views}
\end{figure}

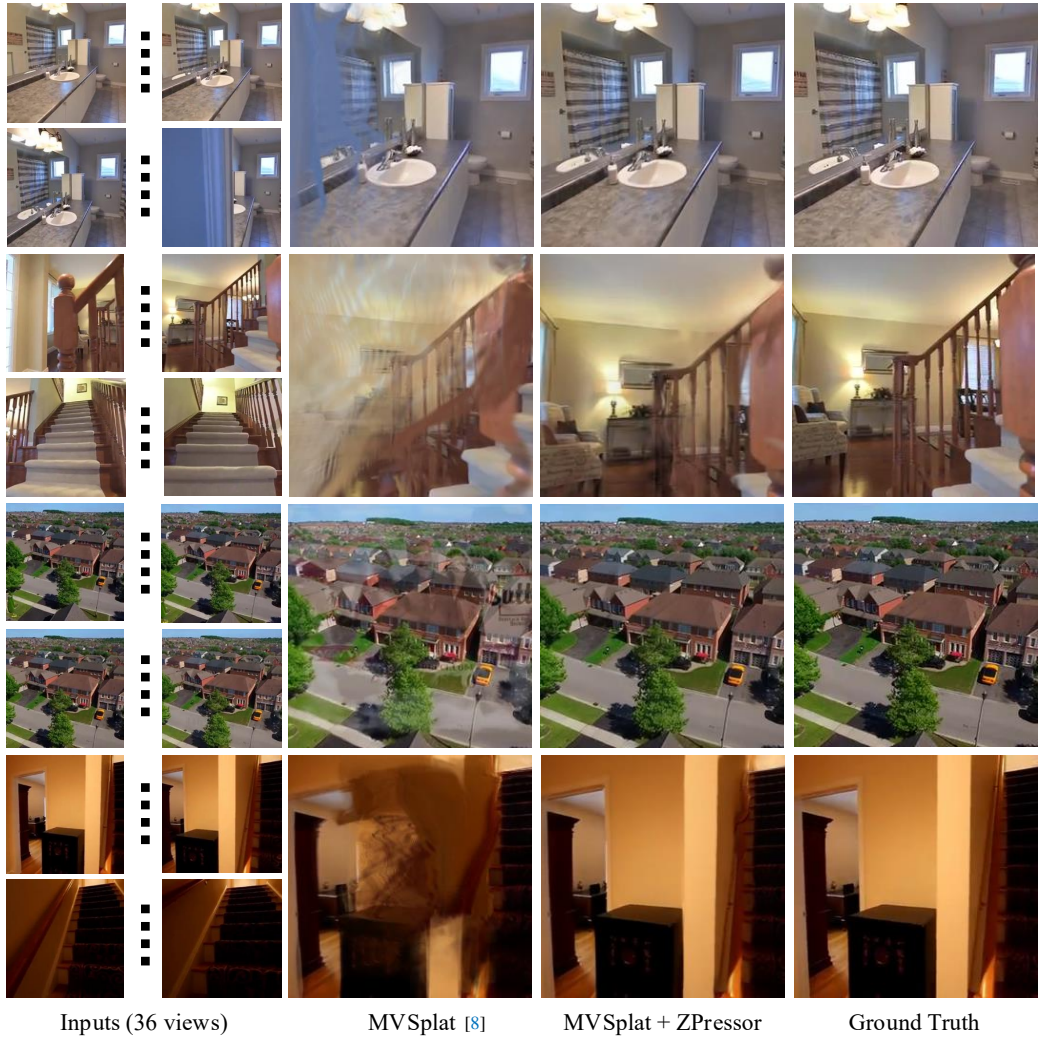
\begin{figure}
    \centering
    \includegraphics[width=\linewidth]{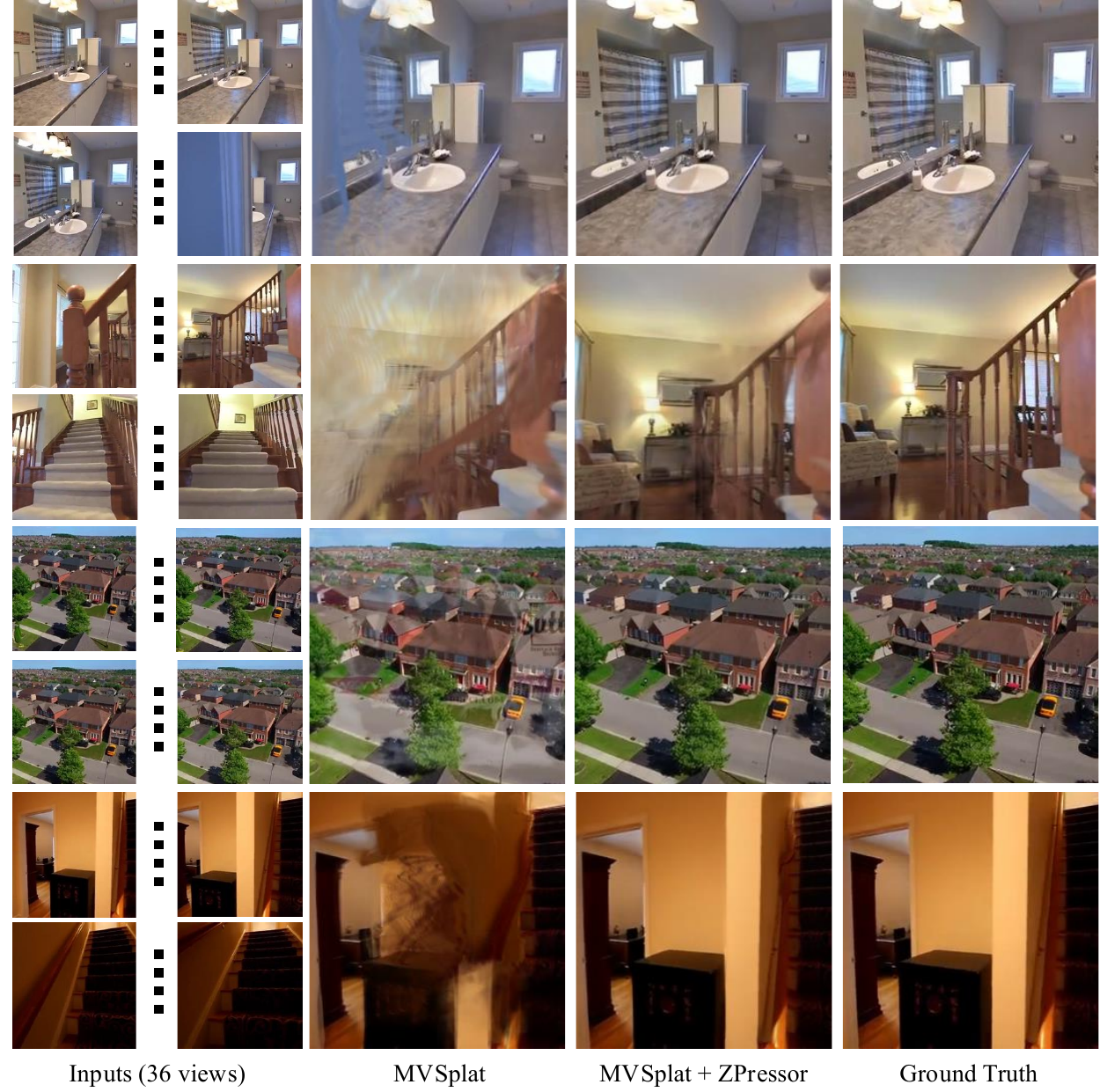}
    \begin{picture}(0,0)
    \put(-21,15){\scriptsize \cite{chen2024mvsplat}}
    \end{picture}
    \caption{\textbf{More qualitative comparisons on RE10K~\cite{zhou2018stereo} with MVSplat~\cite{chen2024mvsplat} under 36 input views.} Models with \method performs the best in all cases.}
    \label{fig:more_re10k_36views}
\end{figure}

This section provides additional qualitative comparison results. We present further visualizations for DepthSplat~\cite{xu2024depthsplat} on the DL3DV~\cite{ling2024dl3dv} and MVSplat~\cite{chen2024mvsplat} on the RE10K~\cite{zhou2018stereo} in \cref{fig:more_dl3dv_36views} and \cref{fig:more_re10k_36views}, with our \method.

\begin{figure}
    \centering
    \includegraphics[width=\linewidth]{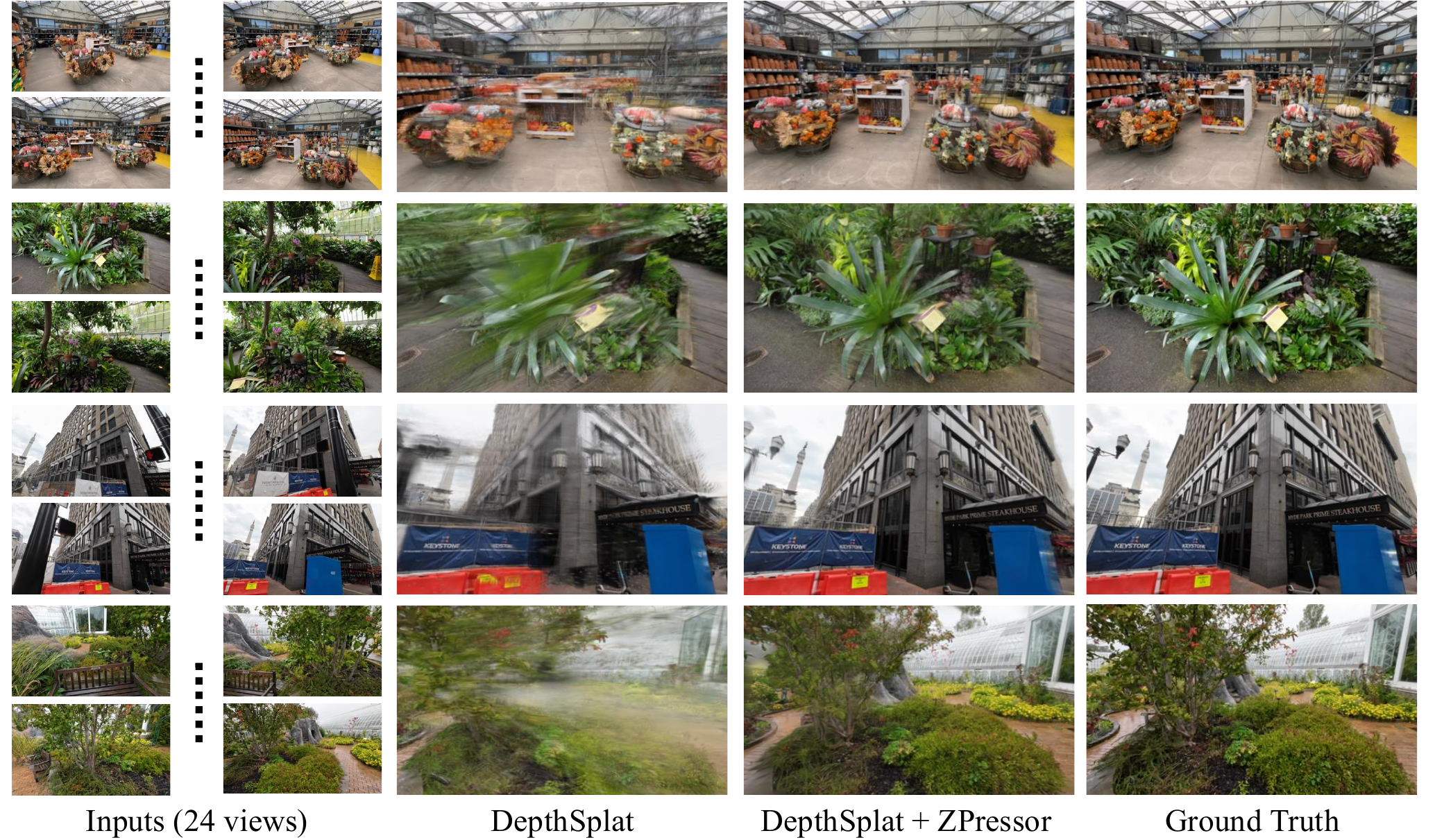}
    \begin{picture}(0,0)
    \put(-20,13.5){\scriptsize \cite{xu2024depthsplat}}
    \end{picture}
    \caption{\textbf{More qualitative comparisons on DL3DV~\cite{ling2024dl3dv} with DepthSplat~\cite{xu2024depthsplat} under 24 input views.}}
    \label{fig:more_dl3dv_24views}
\end{figure}

\begin{figure}
    \centering
    \includegraphics[width=\linewidth]{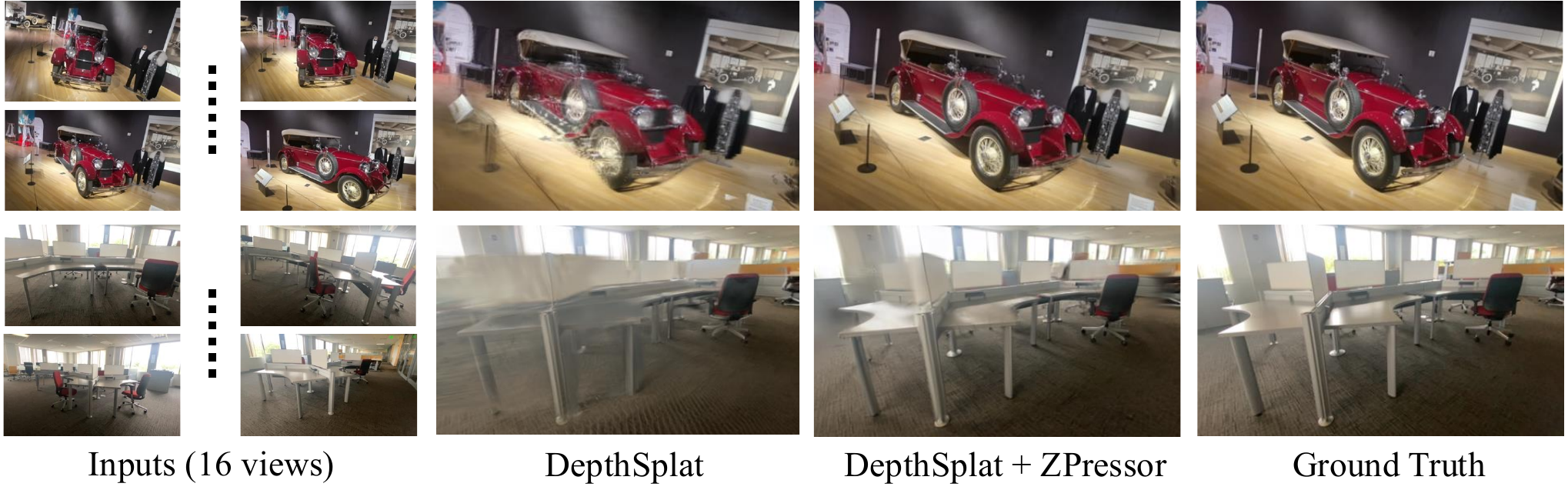}
    \begin{picture}(0,0)
    \put(-18,17.5){\scriptsize \cite{xu2024depthsplat}}
    \end{picture}
    \caption{\textbf{More qualitative comparisons on DL3DV~\cite{ling2024dl3dv} with DepthSplat~\cite{xu2024depthsplat} under 16 input views.}}
    \label{fig:more_dl3dv_16views}
\end{figure}

\begin{figure}
    \centering
    \includegraphics[width=\linewidth]{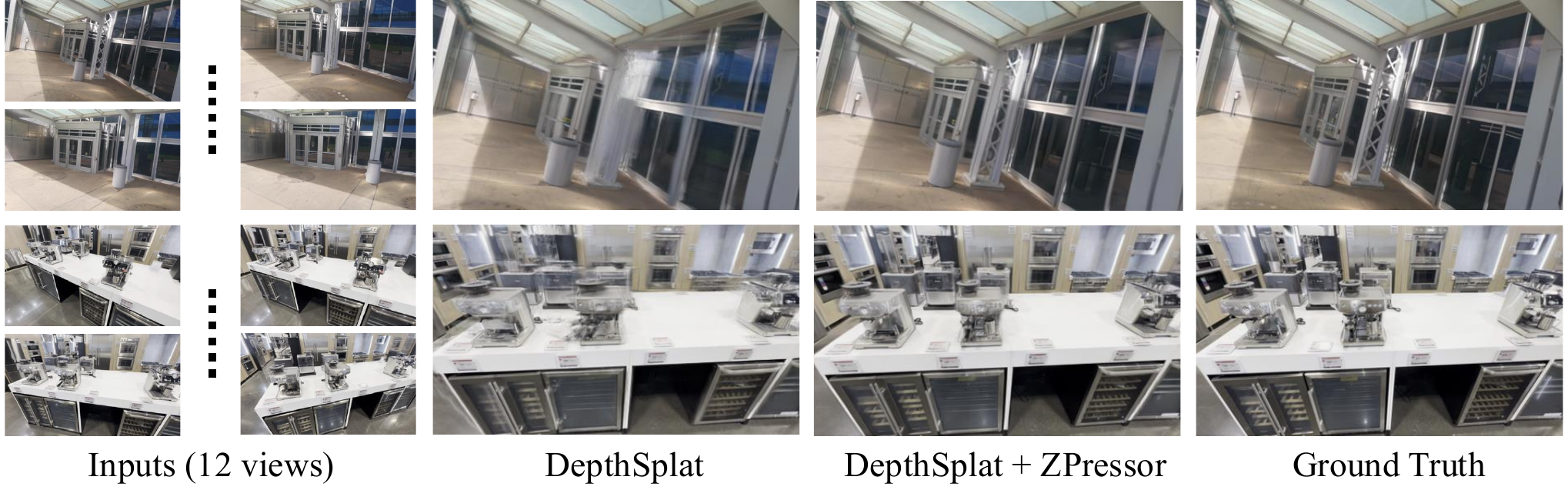}
    \begin{picture}(0,0)
    \put(-18,17.5){\scriptsize \cite{xu2024depthsplat}}
    \end{picture}
    \caption{\textbf{More qualitative comparisons on DL3DV~\cite{ling2024dl3dv} with DepthSplat~\cite{xu2024depthsplat} under 12 input views.}}
    \label{fig:more_dl3dv_12views}
\end{figure}

\begin{figure}
    \centering
    \includegraphics[width=\linewidth]{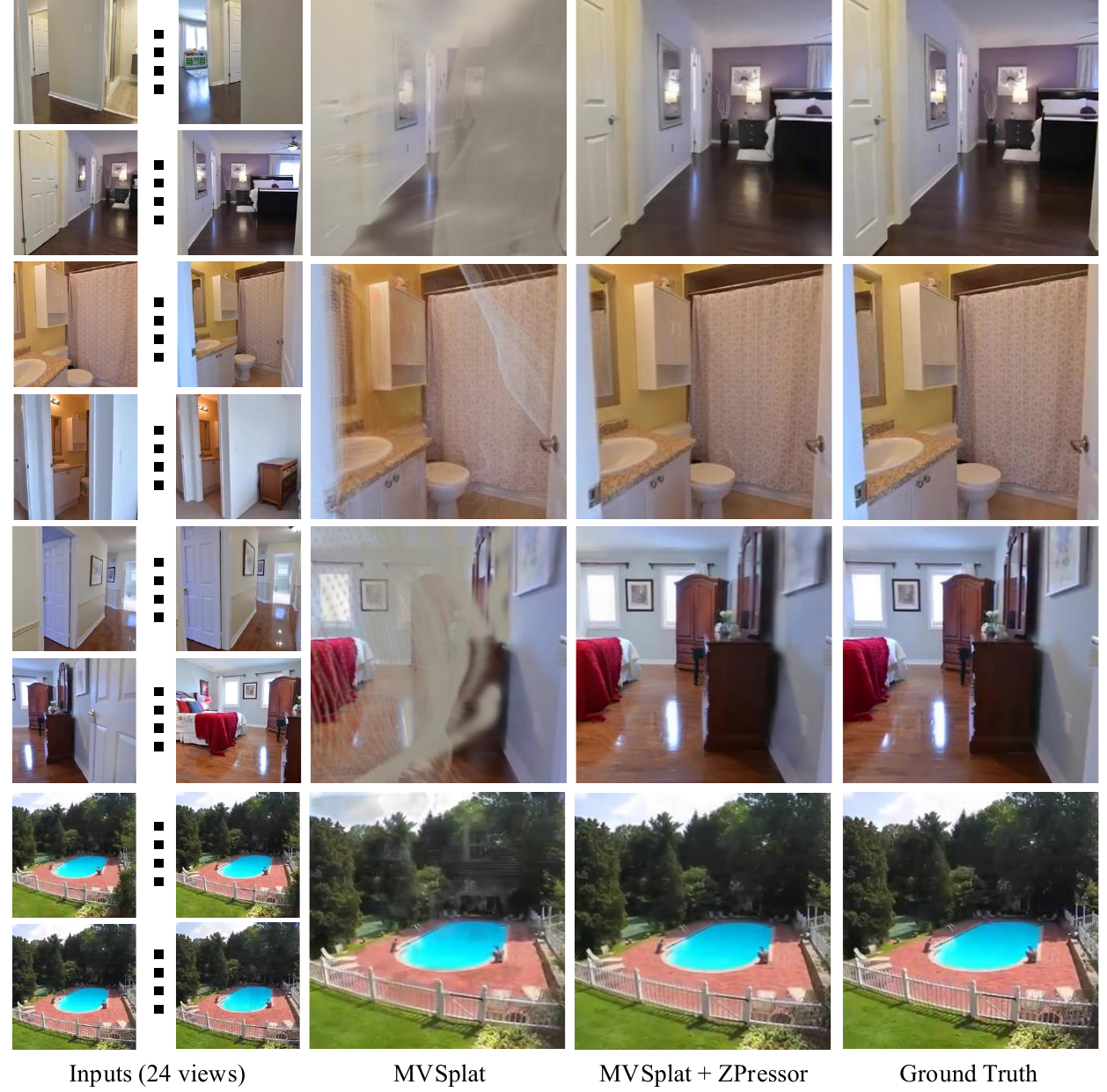}
    \begin{picture}(0,0)
    \put(-21,15){\scriptsize \cite{chen2024mvsplat}}
    \end{picture}
    \caption{\textbf{More qualitative comparisons on RE10K~\cite{zhou2018stereo} with MVSplat~\cite{chen2024mvsplat} under 24 input views.} }
    \label{fig:more_re10k_24views}
\end{figure}

\begin{figure}
    \centering
    \includegraphics[width=\linewidth]{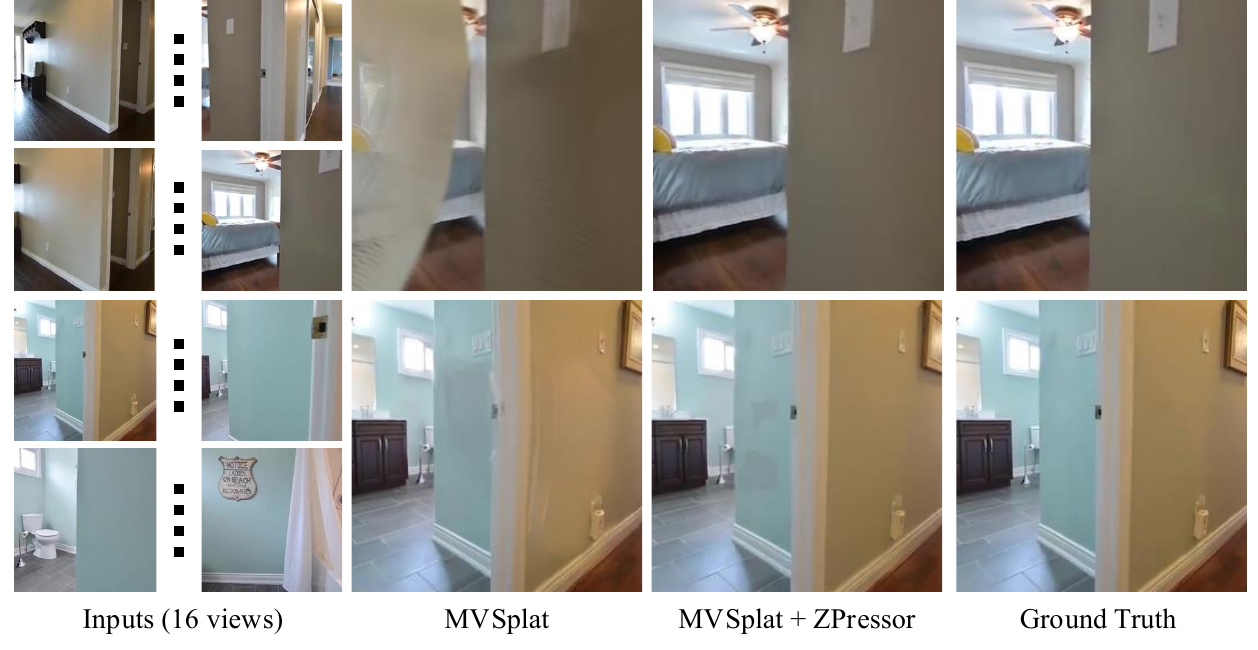}
    \begin{picture}(0,0)
    \put(-22.5,19){\scriptsize \cite{chen2024mvsplat}}
    \end{picture}
    \caption{\textbf{More qualitative comparisons on RE10K~\cite{zhou2018stereo} with MVSplat~\cite{chen2024mvsplat} under 16 input views.} }
    \label{fig:more_re10k_16views}
\end{figure}

\begin{figure}
    \centering
    \includegraphics[width=\linewidth]{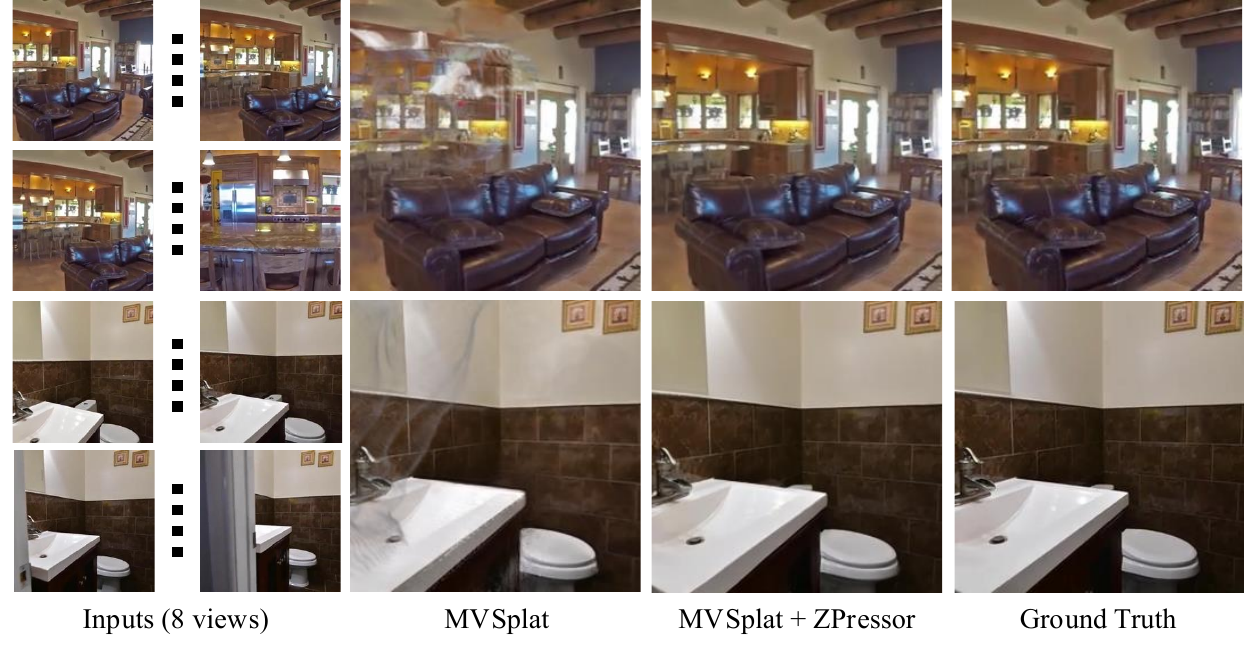}
    \begin{picture}(0,0)
    \put(-22.5,19){\scriptsize \cite{chen2024mvsplat}}
    \end{picture}
    \caption{\textbf{More qualitative comparisons on RE10K~\cite{zhou2018stereo} with MVSplat~\cite{chen2024mvsplat} under 8 input views.} }
    \label{fig:more_re10k_8views}
\end{figure}

Furthermore, to illustrate how \method performs with dense input views, we showcase comparative results. For DepthSplat~\cite{xu2024depthsplat}, comparisons between the original framework and DepthSplat augmented with \method are presented for scenarios with 24, 16, and 12 input views in \cref{fig:more_dl3dv_24views}, \cref{fig:more_dl3dv_16views}, and \cref{fig:more_dl3dv_12views}. Similarly, for MVSplat~\cite{chen2024mvsplat}, visual comparisons between the original framework and MVSplat integrated with \method are displayed for inputs of 24, 16, and 8 views in \cref{fig:more_re10k_24views}, \cref{fig:more_re10k_16views}, and \cref{fig:more_re10k_8views}. The corresponding quantitative results for these multi-view experiments can be found in \cref{tab:multi view results re10k}.

\end{document}